%% file: aistats_paper.tex
\documentclass[twoside]{article}

\usepackage[accepted]{aistats2023}
\usepackage[round]{natbib}
\renewcommand{\bibname}{References}
\renewcommand\refname{References}
\renewcommand{\bibsection}{\subsubsection*{\bibname}}

\usepackage[utf8]{inputenc} 
\usepackage[T1]{fontenc}    

\usepackage{booktabs}       
\usepackage{amsfonts}       
\usepackage{nicefrac}       
\usepackage{microtype}      

\usepackage{graphicx}

\usepackage{url}
\usepackage[symbol]{footmisc}

\usepackage{todonotes}
\usepackage[]{algorithm2e}

\usepackage{algorithm,algorithmic}
\usepackage{amsmath}
\usepackage{bigints}
\usepackage{amssymb}
\usepackage{caption}
\usepackage{subcaption}
\usepackage{amsthm}
\usepackage{bbm}
\usepackage{array,multirow}
\usepackage{wrapfig}
\usepackage[normalem]{ulem}
\usepackage{stmaryrd,scalerel}

\definecolor{solin-colour1}{rgb}{0.2667,0.4471,0.7098} 
\definecolor{solin-colour2}{rgb}{0.1647,0.6706,0.3804} 
\definecolor{solin-colour3}{rgb}{0.8275,0.2627,0.3059} 
\definecolor{solin-colour4}{rgb}{0.5216,0.4392,0.7176} 
\definecolor{solin-colour5}{rgb}{0.8118,0.7255,0.4118} 
\definecolor{solin-colour6}{rgb}{0.2745,0.7176,0.8157} 
\definecolor{solin-colour7}{HTML}{BBBBBB} 
\definecolor{light-colour1}{rgb}{0.6902,0.7686,0.8863} 
\definecolor{light-colour2}{rgb}{0.5451,0.8902,0.6941} 
\definecolor{light-colour3}{rgb}{0.9412,0.7490,0.7647} 
\definecolor{light-colour4}{rgb}{0.8627,0.8392,0.9176} 
\definecolor{light-colour5}{rgb}{0.9569,0.9373,0.8667} 
\definecolor{light-colour6}{rgb}{0.7529,0.9020,0.9373} 
\definecolor{light-colour7}{rgb}{0.8750,0.8750,0.8750} 

\usepackage[english]{babel}
\addto{\captionsenglish}{%
    \renewcommand{\bibname}{References}
}
\usepackage{mdframed}   

\definecolor{mydarkblue}{rgb}{0,0.08,0.45}
\usepackage[colorlinks,citecolor=mydarkblue,urlcolor=mydarkblue,linkcolor=mydarkblue,filecolor=mydarkblue, pdfencoding=auto, psdextra]{hyperref}

\input{math_commands.tex}

\theoremstyle{plain}
\newtheorem{theorem}{Theorem}[section]
\newtheorem{proposition}[theorem]{Proposition}

\newtheorem{corollary}[theorem]{Corollary}
\theoremstyle{definition}
\newtheorem{definition}[theorem]{Definition}

\theoremstyle{remark}

\usepackage[capitalize]{cleveref}

\begin{document}

%
\runningtitle{Learning to Defer to Multiple Experts}

%

\twocolumn[

\aistatstitle{Learning to Defer to Multiple Experts: Consistent Surrogate Losses, \\ Confidence Calibration, and Conformal Ensembles}

\aistatsauthor{ Rajeev Verma$^*$ \And Daniel Barrejón$^*$ \And Eric Nalisnick }

\aistatsaddress{University of Amsterdam \And Universidad Carlos III de Madrid \\ \footnotesize{*Equal contribution, order determined by coin flip} \And University of Amsterdam}]

\begin{abstract}
We study the statistical properties of \textit{learning to defer} (L2D) to multiple experts.  In particular, we address the open problems of deriving a consistent surrogate loss, confidence calibration, and principled ensembling of experts.  Firstly, we derive two consistent surrogates---one based on a softmax parameterization, the other on a one-vs-all (OvA) parameterization---that are analogous to the single expert losses proposed by \citet{pmlr-v119-mozannar20b} and \citet{verma2022calibrated}, respectively.  We then study the frameworks' ability to estimate $\mathbb{P}(\rsm_{j} = \ry | \vx)$, the probability that the $j$th expert will correctly predict the label for $\vx$.  Theory shows the softmax-based loss causes mis-calibration to propagate between the estimates while the OvA-based loss does not (though in practice, we find there are trade offs).  Lastly, we propose a conformal inference technique that chooses a subset of experts to query when the system defers.  We perform empirical validation on tasks for galaxy, skin lesion, and hate speech classification.
\end{abstract}

\section{INTRODUCTION}
\label{sec:intro}
Solving complex problems often requires the involvement of multiple experts \citep{fay2006getting}.  These experts may have non-overlapping specialties, such as in a large construction project that requires the advice of engineers, architects, geologists, lawyers, etc.  Or perhaps the difficulty of the task requires multiple opinions, as when a team of doctors consults on a difficult medical diagnosis.  Thus, modern \textit{hybrid intelligent} (HI) systems \citep{kamar2016directions, dellermann2019hybrid, akata2020research}---so called because they combine computational and human decision making---need to support the participation of multiple experts.

\textit{Learning to defer} (L2D) \citep{10.5555/3327345.3327513} presents an elegant framework for implementing HI systems.  A \textit{rejector} model acts as a meta-classifier, predicting whether the downstream classifier or human is more likely to make the correct decision for a given input.  Yet existing L2D frameworks do not obviously accommodate additional experts.  For instance, the rejector's job becomes more challenging when there are multiple experts. It has two decisions to make: \emph{when} to defer and \emph{to which} expert.  The latter decision is equally important, and thus verifying that the rejector can accurately monitor expert quality is essential for safe and effective deployment.  

In this paper, we develop the statistical foundations of multiclass L2D with multiple experts.  Specifically, we address the following open problems: deriving a consistent surrogate loss, studying whether these systems are confidence calibrated, and developing a principled technique for ensembling expert decisions.  The first and second contributions ensure the soundness of the optimization problem and resulting downstream decision making.  Our third contribution, expert ensembles, follows from our study of calibration, as we propose a conformal inference procedure for selecting a subset of experts.  We empirically validate our methods on the tasks of image classification, galaxy categorization, skin lesion diagnosis, and hate speech detection.  We find that our consistent losses result in superior accuracy and calibration when compared to existing systems based on (inconsistent) mixtures of experts \citep{hemmer2022forming}.   

\section{LEARNING TO DEFER}
\citet{pmlr-v119-mozannar20b} and \citet{verma2022calibrated} proposed the only known consistent (surrogate) loss functions for multiclass L2D.  Thus we focus on their formulations.  We provide the technical background of L2D in this section before moving on to the multi-expert setting. 

\paragraph{Data} We first define the data for multiclass L2D with one expert.  Let $\mathcal{X}$ denote the feature space, and let $\mathcal{Y}$ denote the label space, which we assume to be a categorical encoding of multiple ($K$) classes.  $\rvx_n \in \mathcal{X}$ denotes a feature vector, and $\ry_n \in \mathcal{Y}$ denotes the associated class defined by $\mathcal{Y}$ (1 of $K$).  The L2D problem also assumes that we have access to (human) expert demonstrations.  Denote the expert's prediction space as $\mathcal{M}$, which is usually taken to be equal to the label space: $\mathcal{M}$ = $\mathcal{Y}$.  Expert demonstrations are denoted $\rsm_n \in \mathcal{M}$ for the associated features $\rvx_n$.  The combined N-element training sample is $\mathcal{D} = \{\vx_n, y_n, m_n\}_{n=1}^{N}$.



\paragraph{Models and Learning} \citet{pmlr-v119-mozannar20b}'s and \citet{verma2022calibrated}'s L2D frameworks compose two models: a classifier and a rejector \citep{46544,NIPS2016_7634ea65}.  Denote the \textit{classifier} as $h: \mathcal{X} \rightarrow \mathcal{Y}$ and the \textit{rejector} as $r: \mathcal{X} \rightarrow \{0,1\}$.  When $r(\rvx)=0$, the classifier makes the decision. When $r(\rvx)=1$, the system defers the decision to the human.  When the classifier makes the prediction, then the system incurs a loss $\ell(h(\vx), y)$.  When the human makes the prediction (i.e.~$r(\vx)=1$), the system incurs a loss $\ell_{\text{exp}}(m, y)$.  Using the rejector to combine these losses, we have the overall classifier-rejector loss: 
\begin{equation}\begin{split}
    L(&h, r) = \\ & \mathbb{E}_{\rvx, \ry, \rsm}\left[(1-r(\rvx)) \  \ell(h(\rvx), \ry) \ + \ r(\rvx) \ \ell_{\text{exp}}(\rsm, \ry) \right]
\end{split}
\end{equation} where the rejector is acting as an indicator function that controls which loss to use.  While this formulation is valid for general losses, the canonical $0-1$ loss is of special interest for classification tasks:
\begin{equation}\label{eq:0-1}\begin{split}
 &L_{0-1}(h, r) = \\ & \ \ \ \ \ \mathbb{E}_{\rvx, \ry, \rsm}\left[(1-r(\rvx)) \  \mathbb{I}[h(\rvx) \ne \ry] \ + \ r(\rvx) \ \mathbb{I}[\rsm \ne \ry] \right]
\end{split}
\end{equation} where $\mathbb{I}$ denotes an indicator function that checks if the prediction and label are equal.  Upon minimization, the resulting Bayes optimal classifier and rejector satisfy: 
\begin{equation}\begin{split}\label{eq:Bayes_optimal_rej_clf_0-1_loss}
    h^{*}(\vx) &= \argmax_{y \in \mathcal{Y}} \ \mathbb{P}(\ry = y | \vx), \\
    r^{*}(\vx) &= \mathbb{I}\left[\mathbb{P}(\rsm = \ry | \vx) \ge \max_{y \in \mathcal{Y}} \mathbb{P}(\ry = y | \vx) \right],
\end{split}
\end{equation} where $\mathbb{P}(\ry | \vx)$ is the probability of the label under the data generating process, and $\mathbb{P}(\rsm = \ry | \vx)$ is the probability that the expert is correct.  The expert likely will have additional knowledge not available to the classifier, which possibly allows the expert to outperform the Bayes optimal classifier.

\paragraph{Softmax Surrogate} \citet{pmlr-v119-mozannar20b} proposed the first consistent surrogate loss for $L_{0-1}$, meaning that its minimizers agree with the Bayes optimal solutions in Equation \ref{eq:Bayes_optimal_rej_clf_0-1_loss}.  They accomplish this by first unifying the classifier and rejector via an augmented label space that includes the rejection option.  Formally, this label space is defined as $\mathcal{Y}^{\bot} = \mathcal{Y} \cup \{\bot\}$ where $\bot$ denotes the rejection option.  Secondly, \citet{pmlr-v119-mozannar20b} use a reduction to cost sensitive learning that ultimately resembles the cross-entropy loss for a softmax parameterization.  Let $g_{k}:\mathcal{X} \mapsto \mathbb{R}$ for $k \in [1, K]$ where $k$ denotes the class index, and let $g_{\bot}:\mathcal{X} \mapsto \mathbb{R}$ denote the rejection ($\bot$) option.  These $K+1$ functions are then combined in the following softmax-based, point-wise surrogate loss:
\begin{equation}\label{eq:sm_loss}\begin{split}
     \rphi_{\text{SM}}(g_{1}&,\ldots, g_{K}, g_{\bot}; \vx, y, m) =  \\ & -\log \left( \frac{\exp\{g_{y}(\vx)\}}{\sum_{y' \in \mathcal{Y}^{\bot}} \exp\{g_{y'}(\vx)\} }\right) \\ & - \mathbb{I}[m = y]  \ \log \left( \frac{\exp\{g_{\bot}(\vx)\}}{\sum_{y' \in \mathcal{Y}^{\bot}} \exp\{g_{y'}(\vx)\} }\right).
\end{split}
\end{equation} The intuition is that the first term maximizes the function $g_{k}$ associated with the true label.  The second term then maximizes the rejection function $g_{\bot}$ but only if the expert's prediction is correct.  At test time, the classifier is obtained by taking the maximum over $k \in [1, K]$: $\hat{y} = h(\vx) = \argmax_{k \in [1, K]} g_{k}(\vx)$.  The rejection function is similarly formulated as $r(\vx) = \mathbb{I}[ g_{\bot}(\vx) \ge \max_{k} g_{k}(\vx) ]$.  

\paragraph{One-vs-All Surrogate}
\citet{verma2022calibrated} proposed an alternative consistent surrogate for multiclass L2D based on a one-vs-all (OvA) formulation.  Again assume we have $K+1$ functions $g_{1}(\rvx),\ldots, g_{K}(\rvx), g_{\bot}(\rvx)$ such that $g: \mathcal{X} \mapsto \mathbb{R}$.  Their one-vs-all (OvA) surrogate loss has the point-wise form:
\begin{equation}\label{eq:ova_loss}\begin{split}
     & \rpsi_{\text{OvA}}(g_{1},\ldots, g_{K}, g_{\bot}; \vx, y, m) =  \\ & \ \ \  \rphi[g_{y}(\vx)] + \sum_{y' \in \mathcal{Y}, y' \ne y} \rphi[-g_{y'}(\vx)] \ \ + \\ & \ \ \   \rphi[-g_{\bot}(\vx)] + \mathbb{I}[m=y]\left(\rphi[g_{\bot}(\vx)] - \rphi[-g_{\bot}(\vx)]\right)
\end{split}
\end{equation} where $\rphi:\{\pm 1\}\times \mathbb{R} \mapsto \mathbb{R}_{+}$ is a binary surrogate loss.  For instance, when $\rphi$ is the logistic loss, we have $\rphi[f(\vx)] = \log(1 + \exp\{-f(\vx)\})$.  The classifier and rejector are computed exactly the same as for the softmax loss.  The motivation for this loss is that it produces better calibrated systems than those produced by the softmax-based loss.  The softmax loss has a degenerate parameterization that causes it, in practice, to overestimate the expert’s probability of correctness \citep{verma2022calibrated}.   

\section{L2D TO MULTIPLE EXPERTS}
We now turn to the multi-expert setting, deriving two consistent surrogate losses that are analogous to \citet{pmlr-v119-mozannar20b}'s and \citet{verma2022calibrated}'s single-expert loss functions. 

\paragraph{Data} Again let $\rvx_n \in \mathcal{X}$ and $\ry_n \in \mathcal{Y}$ be the feature and label respectively, as defined above.  Now let there be $J$ experts, and denote each expert's prediction space as $\mathcal{M}_{j}$ (which again we will assume is equal to the label space: $\mathcal{M}_{j}$ = $\mathcal{Y}$ $ \ \forall j$). The expert demonstrations are denoted $\rsm_{n,j} \in \mathcal{M}_{j}$ for the associated features $\rvx_n$.  The combined N-element training sample is then denoted $\mathcal{D} = \{\vx_n, y_n, m_{n,1}, \ldots, m_{n,J}\}_{n=1}^{N}$.


\paragraph{Models} Again we use the classifier-rejector formulation.  The classifier ($h$) is unchanged from the single-expert setting.  The rejector, on the other hand, must be modified.  In L2D with one expert, the rejector makes a binary decision---to defer or not.  In multi-expert L2D, the rejector also must choose \emph{to which} expert to assign the instance.  Hence let the rejector be denoted $r: \mathcal{X} \rightarrow \{0,1,\ldots, J\}$.  When $r(\rvx)=0$, the classifier makes the decision.  When $r(\rvx)=j$, the system defers the decision to the $j$th expert.

\paragraph{Learning} Again the learning objective is the $0-1$ loss. We can re-write Equation \ref{eq:0-1} for the multi-expert setting as:
\begin{equation}\label{eq:0-1_multi}\begin{split}
 &L_{0-1}(h, r) =  \mathbb{E}_{\rvx, \ry, \{\rsm_j\}_{j=1}^{J}}\Bigg[\mathbb{I}[r(\rvx) = 0] \  \mathbb{I}[h(\rvx) \ne \ry] \\ & \ \ \ \ \ \ \ \ \ \ \ \ \ \ \ \ \ \ \ \ \ \ \ \ \ \ \  \ \ \ \ \ \ \ \ + \ \sum_{j=1}^{J} \mathbb{I}[r(\rvx) = j] \ \mathbb{I}[\rsm_{j} \ne \ry] \Bigg]
\end{split}
\end{equation}
The corresponding Bayes optimal classifier and rejector are: 
\begin{equation}\label{bayes_opt_multi}\begin{split}
    h^{*}(\vx) &= \argmax_{y \in \mathcal{Y}} \ \mathbb{P}(\ry = y | \vx), \\
    r^{*}(\vx) &= \begin{cases} &0 \text{ if } \mathbb{P}(\ry = h^{*}(\vx) | \vx) > \mathbb{P}(\rsm_{j'} = \ry | \vx) \ \  \forall j'\\ &\argmax_{j \in [1, J]} \mathbb{P}(\rsm_{j} = \ry | \vx) \ \  \text{ otherwise},
\end{cases}
\end{split}
\end{equation} where $\mathbb{P}(\ry | \vx)$ is again the probability of the label under the data generating process and $\mathbb{P}(\rsm_{j} = \ry | \vx)$ is the true probability that the $j$th expert is correct. We provide the derivation of this rule in Section \ref{sec:Bayes-rule-multiL2D-derivation}.


\subsection{Softmax Surrogate Loss} 
Given the preceding definitions, we can now define the multi-expert analog of the softmax-based surrogate loss.  Define the augmented label space as $\mathcal{Y}^{\bot} = \mathcal{Y} \cup \{\bot_{1}, \ldots, \bot_{J} \}$ where $\bot_{j}$ denotes the decision to defer to the $j$th expert.  Let the classifier be composed of $K$ functions: $g_{k}:\mathcal{X} \mapsto \mathbb{R}$ for $k \in [1, K]$ where $k$ denotes the class index.  Then let the rejector be implemented with $J$ functions: $g_{\bot, j}:\mathcal{X} \mapsto \mathbb{R}$ for $j \in [1, J]$ where $j$ is the expert index.  We propose to combine these $K+J$ functions via the following softmax-parameterized surrogate loss:
\begin{equation}\label{sm_multi_surr}\begin{split}
     &\rphi_{\text{SM}}^{J}\left(g_{1},\ldots, g_{K}, g_{\bot, 1}, \ldots, g_{\bot, J}; \vx, y, m_{1}, \ldots, m_{J}\right) =  \\ & -\log \left( \frac{\exp\{g_{y}(\vx)\}}{\sum_{y' \in \mathcal{Y}^{\bot}} \exp\{g_{y'}(\vx)\} }\right) \\ & - \sum_{j=1}^{J} \mathbb{I}[m_{j} = y]  \ \log \left( \frac{\exp\{g_{\bot,j}(\vx)\}}{\sum_{y' \in \mathcal{Y}^{\bot}} \exp\{g_{y'}(\vx)\} }\right).
\end{split}
\end{equation} The intuition is that the first term maximizes the function $g_{k}$ associated with the true label.  The second term maximizes the rejection function $g_{\bot, j}$ but only if the $j$th expert's prediction is correct.  At test time, the classifier is obtained by taking the maximum over $k \in [1, K]$: $\hat{y} = h(\vx) = \argmax_{k \in [1, K]} g_{k}(\vx)$.  The rejection function is similarly formulated as $$r(\vx) = \begin{cases}
  &0 \ \ \text{ if } \ g_{h(\vx)} > g_{\bot,j'} \ \  \forall j' \in [1, J]\\
  &\argmax_{j \in [1, J]} g_{\bot,j}(\vx) \ \  \text{ otherwise}.
\end{cases}$$ 

Our proof of the soundness of Equation \ref{sm_multi_surr} follows the same approach that \citet{pmlr-v119-mozannar20b} employed---specifically, a reduction to cost-sensitive learning.  
\begin{theorem}
$\rpsi_{\text{SM}}^{J}$ (Equation \ref{sm_multi_surr}) is a convex (in $g$), calibrated surrogate loss for the $0-1$ multi-expert learning to defer loss (Equation \ref{eq:0-1_multi}).
\end{theorem}
The complete proof is in Appendix \ref{sec:proof-theorem-3.1-main-text}.  The result guarantees that the minimizers $g^{*}_{1},\ldots, g^{*}_{K}, g^{*}_{\bot,1}, \ldots, g^{*}_{\bot,J}$ correspond to the Bayes optimal classifier and rejector given in Equation \ref{bayes_opt_multi}.

\subsection{One-vs-All Surrogate Loss}
We next turn to the OvA surrogate loss.  Let the label space $\mathcal{Y}^{\bot}$ and the functions $g_{1},\ldots, g_{K}, g_{\bot,1}, \ldots, g_{\bot,J}$ be defined just as above for the softmax case.  The OvA-based multi-expert L2D surrogate is then:
\begin{equation}\label{eq:ova_loss_multi}\begin{split}
     & \rpsi_{\text{OvA}}^{J}(g_{1},\ldots, g_{K}, g_{\bot,1},\ldots, g_{\bot,J}; \vx, y, m_{1},\ldots, m_{J}) =  \\ & \rphi[g_{y}(\vx)] + \sum_{y' \in \mathcal{Y}, y' \ne y} \rphi[-g_{y'}(\vx)] \ \ + \sum_{j=1}^{J} \rphi[-g_{\bot,j}(\vx)] \\ & + \sum_{j=1}^{J} \mathbb{I}[m_{j}=y]\left(\rphi[g_{\bot,j}(\vx)] - \rphi[-g_{\bot,j}(\vx)]\right)
\end{split}
\end{equation} where $\rphi:\{\pm 1\}\times \mathbb{R} \mapsto \mathbb{R}_{+}$ is again a binary surrogate loss.  The classifier and rejector are computed exactly as in the softmax case.

We cannot construct our consistency proof in the same direct manner used in the softmax case.  Like \citet{verma2022calibrated}, we proceed by the method of \textit{error correcting output codes} \citep{10.5555/1622826.1622834, Langford05sensitiveerror, 10.1162/15324430152733133, pmlr-v35-ramaswamy14}, a general technique for reducing multiclass problems to multiple binary problems.  We prove the consistency of $\rpsi_{\text{OvA}}^{J}$ by way of the following two results. \begin{theorem}\label{thm: OvA_loss_calibration_multi}
For a strictly proper binary composite loss $\rphi$ with a well-defined continuous inverse link function $\rgamma^{-1}$, $\rpsi_{\text{OvA}}^{J}$ (Equation \ref{eq:ova_loss_multi}) is a calibrated surrogate for the $0-1$ multi-expert learning to defer loss (Equation \ref{eq:0-1_multi}).
\end{theorem}
The complete proof is in Appendix \ref{sec:proof-theorem-3.2-main-text}.  Assuming \textit{minimizability} \citep{Steinwart2007HowTC}---i.e.~that our hypothesis class is sufficiently large (all measurable functions)---the calibration result from Theorem \ref{thm: OvA_loss_calibration_multi} implies consistency. \begin{corollary}\label{cor:OvA_loss_consistent}
Assume that $g \in \mathcal{F}$, where $\mathcal{F}$ is the hypothesis class of all measurable functions.  \textit{Minimizability} \citep{Steinwart2007HowTC} is then satisfied for $\rpsi_{\text{OvA}}^{J}$, and it follows that $\rpsi_{\text{OvA}}^{J}$ is a consistent surrogate for the $0-1$ multi-expert learning to defer loss (Equation \ref{eq:0-1_multi}).
\end{corollary} 
Thus, the minimizers of $\rpsi_{\text{OvA}}^{J}$ (over all measurable functions) agree with the Bayes optimal classifier and rejector (Equation \ref{bayes_opt_multi}).

\subsection{Inconsistency of Mixture of Experts} \label{sec:moe}
While we are the first to propose a consistent surrogate loss, previous work has proposed a \textit{mixture of experts} (MoE) approach to multi-expert L2D.  \citet{hemmer2022forming} formulated the following model of the probability of the label under the whole team (of $J$ experts and one classifier):
\begin{equation*}\begin{split}
p&\left(\ry | \rvx, \rsm_{1},\ldots,\rsm_{J}; \vtheta_{w}, \vtheta_{h} \right) = \\ & w_{0}(\rvx; \vtheta_{w}) \cdot p(\ry | \rvx; \vtheta_{h}) +  \sum_{j=1}^{J} \mathbb{I}[\rsm_{j}=\ry]\cdot w_{j}(\rvx; \vtheta_{w})
\end{split}
\end{equation*} 
where $p(\ry | \rvx; \vtheta_{h})$ denotes the classifier's probability.  The function $\rvw(\rvx; \vtheta_{w}) \in \boldsymbol{\triangle}^{J+1}$---where $\boldsymbol{\triangle}^{J+1}$ is the $(J+1)$-dimensional simplex---defines the mixture weights.  $w_{0}$ assigns weight to the classifier, and $w_{j}$ for $j \in [1,J]$ denotes the weight given to the $j$th expert.  At test time, the index of the maximum weight determines to which downstream decision maker to assign responsibility.  \citet{hemmer2022forming} fit this MoE model using the negative log-likelihood of $p(\ry | \rvx, \rsm_{1},\ldots,\rsm_{J}; \vtheta_{w}, \vtheta_{h})$; denote their loss $L_{\text{MoE}}(\vtheta_{w}, \vtheta_{h})$.  In Appendix \ref{prop:moe_inconsis}, we show that $L_{\text{MoE}}(\vtheta_{w}, \vtheta_{h})$ is inconsistent. We provide a full discussion of related work in Section \ref{sec:related_work}.   



\section{CONFIDENCE CALIBRATION}\label{sec:confidence-calibration-theory-section}
We next turn to the \textit{calibration} \citep{dawid1982well} properties of multi-expert L2D.  While training with a consistent loss should produce models that are well-calibrated, previous work on the single-expert setting found that the underlying parameterizations can strongly influence calibration in practice.  Specifically, \citet{verma2022calibrated} show that the softmax formulation's estimators can be unbounded, resulting in `probability' estimates above one.  As for the calibration of the classifier, \citet{verma2022calibrated} found that there is to systemic issue and can be improved with standard post-hoc techniques like temperature scaling \citep{kull2019temperature}, if necessary.  Their findings also apply to the multi-expert scenario, and thus we consider only the rejector going forward.  

We are particularly interested in the rejector's ability to estimate $\mathbb{P}(\rsm_{j} = \ry | \vx)$, the conditional probability that the $j$th expert is correct.  If the L2D system says that $\mathbb{P}(\rsm_{j} = \ry | \vx_{0}) = 0.7$, then the $j$th expert should be correct $70\%$ of the time for inputs very similar to $\vx_{0}$.   This quantity is crucial not only for the system's ability to correctly defer but is also useful for interpretability and safety---to quantify what the model thinks that the human knows.  

We next define the relevant notion of calibration.  For an estimator of expert correctness $t(\vx): \mathcal{X} \mapsto (0, 1)$, we call $t$ \textit{calibrated} if, for any confidence level $c \in (0, 1)$, the actual proportion of times the expert is correct is equal to $c$: \begin{equation}\begin{split}
     \mathbb{P}(\rsm = \ry \ | \  t(\vx) = c) = c.
\end{split}
\end{equation} This statement should hold for all possible instances $\vx$ with confidence $c$.  Since expert correctness is a binary classification problem, distribution calibration, confidence calibration, and classwise calibration all coincide \citep{pmlr-v89-vaicenavicius19a}.  We can measure the degree of calibration using \textit{expected calibration error} (ECE).  In this case, the relevant ECE is defined as $$\text{ECE}(t) = \mathbb{E}_{\rvx}|\mathbb{P}\left( \rsm = \ry \  | \ t(\rvx) = c \right) - c|,$$ where $\mathbb{E}_{\rvx}$ is usually approximated with samples.

\subsection{Softmax Parameterization} 
For the softmax formulation, the estimator of the probability that the $j$th expert is correct can be derived as follows; see Appendix \ref{sec:proof-theorem-3.1-main-text} (Equation \ref{eqn:softmax-bayes-rejector-condition}).  The Bayes optimal functions $g^{*}_{1},\ldots,g^{*}_{\bot, J}$ have the following relationship with the underlying probability of expert correctness: 
\begin{equation}\label{eq:expert_prob_softmax1}\begin{split}
     &\frac{\mathbb{P}(\rsm_{j} = \ry | \vx)}{1 + \sum_{j' = 1}^{J}\mathbb{P}(\rsm_{j'} = \ry | \vx)} = \\  & \ \ \ \ \ \ \ \ \ \ \ \ \ \ \ \ \ \ \ \ \ \ \ \ \ \ \ \ \ \ \ \ \underbrace{\frac{\exp\{g_{\bot,j}^{*}\}}{\sum_{y' \in \mathcal{Y}^{\bot}}\exp\{g^{*}_{y'}(\vx)\}}}_{t_{\bot, j}^{*}(\vx)}.
\end{split}
\end{equation} 
Denote the RHS of Equation~\ref{eq:expert_prob_softmax1} as $t_{\bot, j}^{*}(\vx)$. Since we have $J$ equations, one for each expert, we can uniquely solve for $\mathbb{P}(\rsm_{j} = \ry \vert \vx)$ as:
\begin{equation}\label{eq:expert_prob_softmax2}
  \mathbb{P}(\rsm_{j} = \ry \vert \vx) = \frac{t_{\bot, j}^{*}(\vx)}{1 - \sum_{j'=1}^{J}t_{\bot, j'}^{*}(\vx)}.  
\end{equation}  Equation \ref{eq:expert_prob_softmax2} exhibits the same pathology as the single expert setting: it is unbounded from above.  For $t_{\bot, j}(\vx) > 0$, as $\sum_{j'=1}^{J}t_{\bot, j'}(\vx)$ approaches one, the estimate of $\mathbb{P}(\rsm_{j} = \ry \vert \vx)$ will go to infinity.  Moreover, the estimator for the $j$th expert depends on the estimators for the other experts.  Thus, if one $t_{\bot, j}(\vx)$ is mis-calibrated, this error will likely propagate to the other estimators.

\subsection{One-vs-All Parameterization}
For the OvA formulation, the probability that the expert is correct is directly modeled by the $j$th deferral function. For the logistic binary loss $\rphi$, we have: \begin{equation}\label{eq:expert_prob_ova}\begin{split}
    \mathbb{P}(\rsm_{j} = \ry | \vx) \  &=  \ \rphi\left(g^{*}_{\bot,j}(\vx)\right) \\ & = \ \frac{1}{1 + \exp\{-g^{*}_{\bot,j}(\vx)\}}.
\end{split}
\end{equation}  This estimator has the correct range of $(0, 1)$ for any setting of $g_{\bot,j} \in \mathbb{R}$.  Moreover, there is no dependence across expert deferral functions $g_{\bot,1},\ldots, g_{\bot,J}$.  We expect these properties to result in better calibration in practice.

\section{ENSEMBLING EXPERTS WITH CONFORMAL INFERENCE}\label{sec:conformal-ensemble-theory-statistics}

Multi-expert L2D, as defined above, operates by selecting just one expert upon deferral.  This approach is sensible if querying each expert results in an independent expense (such as a consulting fee).  However, in other settings, the cost incurred by deferring may just be that of time and efficiency (i.e.~a lack of automation).  In this case, the cost of querying additional experts would be negligible; for example, we could send multiple experts simultaneous messages asking for their decisions.  Given the estimators of $\mathbb{P}(\rsm_{j} = \ry | \vx)$ presented in the previous section, it is then natural to ask how we might ensemble experts according to these estimates of correctness.  Below we present a methodology based on \textit{conformal inference} for obtaining dynamic, minimal ensembles of experts.

\paragraph{Conformal Inference} \textit{Conformal inference} (CI) \citep{shafer2008tutorial} constructs a confidence interval (or set) for predictive inference.  In the traditional multiclass classification setting, given a new observation $\vx_{n+1}$, we wish to determine the correct associated label $\ry_{n+1} = y_{n+1}^{*}$, where $y_{n+1}^{*}$ denotes the true class label.  CI allows us to construct a distribution-free confidence set $\mathsf{C}(\vx_{n+1})$ that will cover the true label with \emph{marginal} probability $1-\alpha$:
\begin{equation*}
    \mathbb{P}\left(y_{n+1}^{*} \notin \mathsf{C}(\vx_{n+1}) \right) \ \le \ \alpha \ \ \ \  \forall \ \mathbb{P} \in \mathfrak{P}
\end{equation*} where $\mathfrak{P}$ represents the space of all distributions---hence the `distribution-free' quality.  Denote the test statistic as $S(\vx, y; \mathcal{D})$.  It is known as a \textit{non-conformity} function: a higher value of $S$ means that $(\vx, y)$ is less conforming to the distribution represented (empirically) by $\mathcal{D}$.  Despite this guarantee, CI is only as good as its test statistic in practice.  For instance, the marginal coverage is naively satisfied if we construct the set randomly by setting $\mathsf{C}(\vx) = \mathcal{Y}$ with probability $1-\alpha$ and returning the empty set otherwise.  CI is implemented by calculating the non-conformity function on a validation set and computing the empirical $1-\alpha$ quantile (with a finite sample correction).  At test time, elements are added to the set until the non-conformity function passes the previously-computed quantile.

\paragraph{Conformal Sets of Experts} We propose applying CI to perform uncertainty quantification for the experts.  Thus, here, $\mathsf{C}(\vx)$ represents a set of experts.  Firstly, we assume there is a best expert: for a new observation $\vx_{n+1}$, let $j_{n+1}^{*}$ denote the best expert such that $$\mathbb{P}(\rsm_{j_{n+1}^{*}} = \ry \vert \vx_{n+1}) > \mathbb{P}(\rsm_{e} = \ry \vert \vx_{n+1}) \ \forall e \ne j^{*}_{n+1}.$$  We would then like to construct a set such that $j_{n+1}^{*}$ is covered with marginal probability $1-\alpha$: \begin{equation*}
    \mathbb{P}\left(j_{n+1}^{*} \notin \mathsf{C}(\vx_{n+1}) \right) \ \le \ \alpha \ \ \ \  \forall \ \mathbb{P} \in \mathfrak{P}
\end{equation*} where $\mathsf{C}(\vx_{n+1})$ again denotes the conformal set and $\mathfrak{P}$ is the same as above.  The set will have a dynamic size that changes with $\vx$, ensuring our ensemble makes efficient use of expert queries.  Unlike in most applications of CI, we can use the procedure to form an ensemble by aggregating the predictions of all experts in the set.

\paragraph{Naive Statistic} We start by adapting a score function from multiclass classification. Let $s_{j}(\vx)$ denote the estimator that the $j$th expert is correct.  For the softmax case, $s_{j}(\vx) = t_{\bot, j}(\vx) / (1 - \sum_{j'} t_{\bot, j'}(\vx))$ (Equation \ref{eq:expert_prob_softmax2}), and for OvA, $s_{j}(\vx) = \rphi(g_{\bot,j}(\vx))$ (Equation \ref{eq:expert_prob_ova}).  Let $\pi_{1},\ldots,\pi_{J}$ denote the indices for a descending ordering of the estimators $s_{j}(\vx)$, i.e.~$s_{\pi_{1}}$ is the expert who has the best chance of being correct (according to the rejector).  The resulting non-conformity function and test statistic are:
\begin{equation}
    S\left(\vx, y, m_{1},\ldots, m_{J}; \mathcal{D}\right) = \sum_{e=1}^{E} s_{\pi_{e}}(\vx)
\end{equation} where $\pi_{E}$ is the index of the expert who has the lowest score $s_{\pi_{E}}$ of all \textit{correct experts} ($m = y$).  This expression means that we will keep adding the correctness scores $s$ in descending order until we include all experts who correctly predict the given instance.  Hence $E=J$ only when all experts are correct and $E<J$ otherwise.  


\paragraph{Regularized Statistic} A problem with the statistic above is that multiple experts can be correct, resulting in noise that obscures the identity of the best expert.  In the experiments, we show that this `naive' statistic is not robust to noise, resulting in inflated set sizes (which are sometimes vacuous).  Similar problems are discussed by \citet{angelopoulos2020uncertainty}.  To address this issue, we employ \textit{conformal risk control} \citep{angelopoulos2022conformal}  to directly control the false negative rate.  We create regularized prediction sets as follows:
\begin{equation}\label{eqn:conformal-risk-control}
    C_{\lambda_{\alpha}}(\vx) = \left\{ j: s_{j}(\vx) + \beta\left(s_{j}(\vx) - \kappa \right)> 1 - \lambda_{\alpha} \right\}
\end{equation} where $\beta$ and $\kappa$ are the parameters of the regularization and $\lambda_{\alpha}$ is chosen to have $1-\alpha$ coverage guarantees. In Appendix \ref{sec:ci_hypers}, we describe $\lambda_{\alpha}$ and how we choose the regularization parameters.  The general idea is to choose $\kappa$ so that confidences lower than this threshold can happen with probability at most $\alpha$.  We choose $\beta$ to optimize the size of the sets.



\begin{figure*}[ht]
       \begin{subfigure}[b]{0.305\textwidth}
         \centering
         \includegraphics[width=0.99\linewidth]{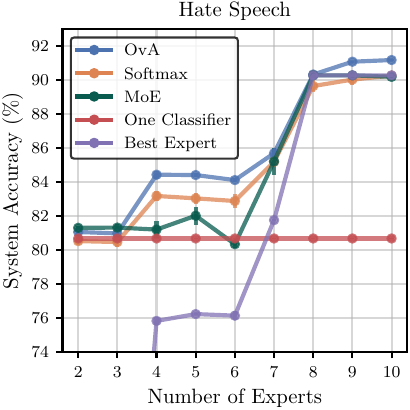}
         \label{fig:hatespeech_sys_acc}
     \end{subfigure}
     \hfill
     \begin{subfigure}[b]{0.305\textwidth}
     \centering
         \includegraphics[width=0.99\linewidth]{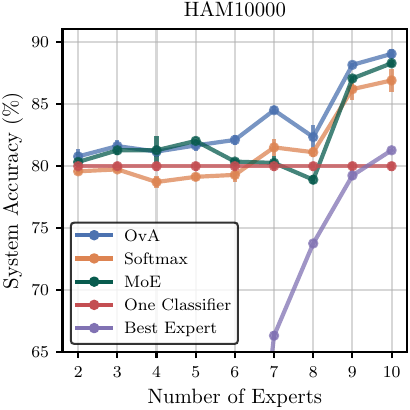}
         \label{fig:ham10000_sys_acc}
     \end{subfigure}
     \hfill
     \begin{subfigure}[b]{0.305\textwidth}
     \centering
         \includegraphics[width=0.99\linewidth]{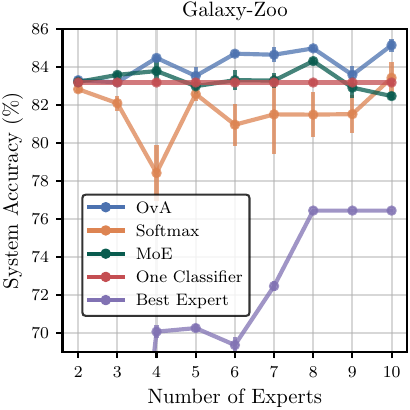}
         \label{fig:galaxy_zoo_sys_acc}
     \end{subfigure}
     \\ 
    \begin{subfigure}[b]{0.305\textwidth}
         \centering
         \includegraphics[width=0.99\linewidth]{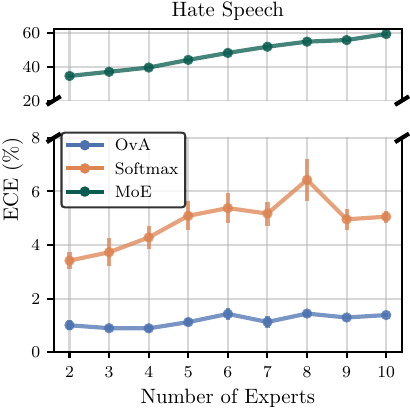}
         \label{fig:hatespeech_ece}
     \end{subfigure}
     \hfill
     \begin{subfigure}[b]{0.305\textwidth}
         \centering
         \includegraphics[width=0.99\linewidth]{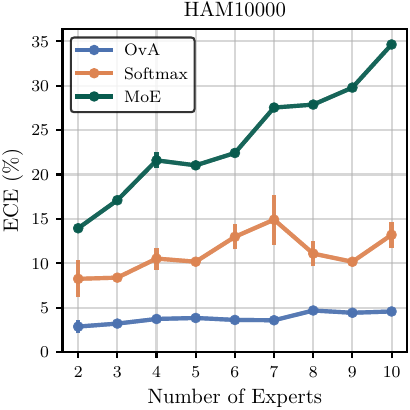}
         \label{fig:ham10000_ece}
     \end{subfigure}
     \hfill
     \begin{subfigure}[b]{0.305\textwidth}
         \centering
         \includegraphics[width=0.99\linewidth]{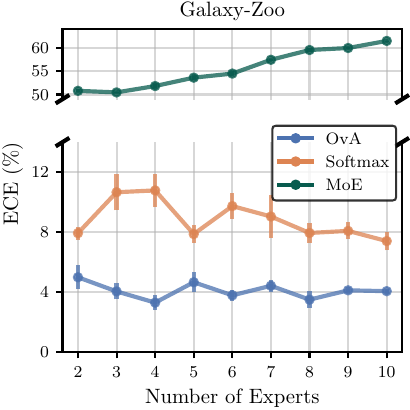}
         \label{fig:galaxy_zoo_ece}
     \end{subfigure}
     \caption{\textit{System Accuracy and Calibration.}  The figures above report the system accuracy (top row) and calibration error (bottom row) as experts of increasing ability are added (from $2$ to $10$).  One classifier (red), best expert (purple), and a mixture of experts (green) \citep{hemmer2022forming} serve as baselines.  We see that the OvA-trained model (blue) performs well in every case.  On the other hand, the softmax-trained model (orange) falls below the one classifier baseline for both \texttt{HAM10000} and \texttt{GalaxyZoo}.} 
     \label{fig:MoE_comparison}
\end{figure*}


\section{RELATED WORK}\label{sec:related_work}
\paragraph{Learning to Defer} \textit{Learning to defer} (L2D) was proposed in its modern form by \citet{10.5555/3327345.3327513}.  Yet classifiers with an option to reject or abstain date back at least to \citet{Chow1957AnOC}.  There have been two primary approaches to making the rejection decision: confidence-based \citep{10.5555/1390681.1442792, JMLR:v11:yuan10a, 10.5555/3327345.3327458, NIPS2008_3df1d4b9, Ramaswamy2018ConsistentAF, Ni2019OnTC} and model-based \citep{46544,NIPS2016_7634ea65}.  Our work, due to it using a parameterized rejector, falls into the latter.  The theoretical properties of the classifier-rejector approach have been well-studied for binary classification \citep{46544, NIPS2016_7634ea65}.  The theory for multiclass classification was first developed by \citet{Ni2019OnTC} and \citet{pmlr-v139-charoenphakdee21a}.  \citet{pmlr-v119-mozannar20b} then built upon this work to establish the first consistent surrogate loss for multiclass L2D.  Previous L2D extensions did not come with consistency guarantees \citep{raghu2019algorithmic, Wilder2020LearningTC,  pradier2021preferential, okati2021differentiable, liu2022incorporating}.  \citet{verma2022calibrated} proposed the second provably consistent surrogate for multiclass L2D based on a one-vs-all formulation. \citet{charusaie2022sample} further studied the L2D optimization problem, proving results for complementarity and active learning.  Our work extends \citet{pmlr-v119-mozannar20b}'s and \citet{verma2022calibrated}'s results to the multi-expert setting---for which no one has yet to propose a consistent surrogate loss. 

\paragraph{Calibration in L2D} \citet{verma2022calibrated} motivate their OvA surrogate from the standpoint of calibration and thus is the only other work that studies the confidence calibration of L2D systems.  We extend their work to the multi-expert setting.  Calibration has received much attention of late in the wider machine learning literature \citep{10.5555/3305381.3305518, kull2019temperature, pmlr-v89-vaicenavicius19a, Gupta2021ToplabelC}.  The dominant methodology is to apply post-hoc calibration: fitting additional parameters on validation data to re-calibrate the formerly mis-calibrated model.  These methods could potentially be applied here---such as, by adding a temperature parameter to the per-expert terms in the OvA loss---but we are primarily interested in the native, `out-of-the-box' calibration properties of the losses.       

\paragraph{Multi-Expert Models} There have been several works that use models to improve the decision making of multiple experts \citep{benz2022counterfactual, straitouri2022provably} and to fuse decisions from models and humans \citep{keswani2021towards, DBLP:journals/corr/abs-2109-14591}.  As mentioned in Section \ref{sec:moe}, \citet{hemmer2022forming} proposed the only existing model for multi-expert L2D.  Yet their approach does not have any supporting theoretical guarantees, such as consistency (like ours). 
 \citet{keswani2021towards} also proposed an MoE-based model but not for the standard L2D setting that we consider.  Rather they allow for responsibility to be passed to multiple downstream sources---specifically, to any of the $2^{J+1}$ possible sets involving the experts and/or model.



\paragraph{Conformal Inference for Human-AI Collaboration}  We know of two works that have used CI for some form of human-AI collaboration.  \citet{straitouri2022provably} apply CI to a classifier and then pass the prediction set to a human to make the final decision. \citet{babbar2022utility} study a similar work flow (apply CI then pass to a human) and also propose applying CI only to non-deferred samples, which results in smaller set sizes.  No previous work has applied CI to obtain sets of experts.

\section{EXPERIMENTS}
Our experimental setup closely follows that of \citet{verma2022calibrated}---but extended to multiple experts.  For all runs, we report the mean and standard error across $3$ random seeds.  We perform three types of experiments.  In the first, we check the system accuracy of the derived consistent surrogate losses in three consequential tasks (Subsection \ref{sec:system-accuracy}): galaxy classification, skin lesion diagnosis, and hate speech detection.  We find that the OvA-trained model often outperforms both the softmax variant and MoE baseline.  Secondly, we investigate the confidence calibration properties of the surrogates (Subsection \ref{sec: calibration}).  As hypothesized, the OvA loss results in less calibration error on both simulated and real data (possibly explaining its superior accuracy).  Lastly, we investigate the efficacy of our conformal ensembling procedure (Subsection \ref{sec:ensemble_results}).  For the naive statistic, the OvA loss' superior calibration results in appropriately smaller sets.  For the regularized statistic, both losses perform equally well. Our implementations are publicly available at \href{https://github.com/rajevv/Multi_L2D}{https://github.com/rajevv/Multi\_L2D}.

%

\begin{figure*}
     \centering
       \begin{subfigure}[b]{0.24\textwidth}
         \centering
         \includegraphics[width=\linewidth]{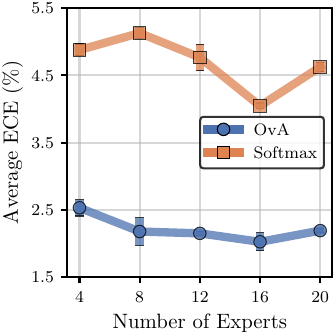}
         \caption{ECE as Experts Increase}
         \label{fig:avg-ECE-increase-experts-CIFAR-10}
     \end{subfigure}
     \hfill
     \begin{subfigure}[b]{0.24\textwidth}
         \centering
         \includegraphics[width=\linewidth]{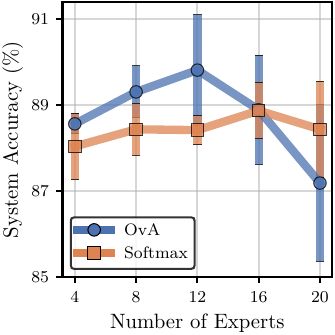}
         \caption{Acc.~as Experts Increase}
         \label{fig:acc-increase-experts-CIFAR-10}
     \end{subfigure}
     \hfill
     \begin{subfigure}[b]{0.24\textwidth}
         \centering
         \includegraphics[width=\linewidth]{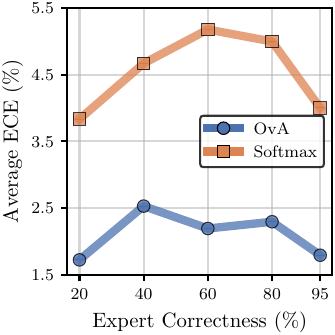}
         \caption{ECE for Improving Experts}
         \label{fig:avg-ECE-increase-confidence-CIFAR-10}
     \end{subfigure}
     \hfill
     \begin{subfigure}[b]{0.24\textwidth}
         \centering
         \includegraphics[width=\linewidth]{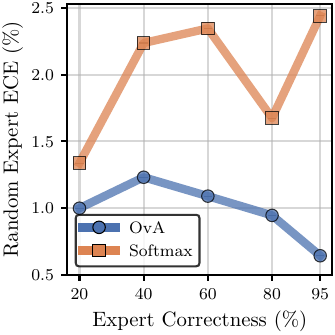}
         \caption{ECE of Random Expert}
         \label{fig:randomexpert-ECE-increase-confidence-CIFAR-10}
     \end{subfigure} 
     \label{fig:cifar-10-calibration-MultiL2D}
     \caption{\textit{Confidence Calibration Simulations.} The figures above report the results for confidence calibration simulations performed on \texttt{CIFAR-10}.  The first and second subfigures show calibration error (a) and system accuracy (b) as the number of experts increases from $4$ to $20$.  We see that the OvA formulation (blue) has better calibration across all runs, but this translate to better accuracy only for $16$ or fewer experts.  The third and forth subfigures show calibration error as experts' abilities increase (a) and when one expert is kept at random chance (b).  OvA (blue) shows better calibration in both metrics.}
\end{figure*}

\subsection{Overall System Accuracy}\label{sec:system-accuracy}
\paragraph{Data Sets} We report the overall system accuracy for three real-world data sets: \texttt{HAM10000} \citep{tschandl2018ham10000} for skin lesions diagnosis, \texttt{Galaxy-Zoo} \citep{10.1111/j.1365-2966.2008.14252.x} for galaxy classification, and \texttt{HateSpeech} \citep{Davidson2017AutomatedHS} for hate speech detection. The train-validation-test split is $60\% - 20\% - 20\%$.  Following \citet{verma2022calibrated}, we down-sample \texttt{Galaxy-Zoo} to $10,000$ instances.

\paragraph{Models} We use a $34$-layer residual network (ResNet34) and a $50$-layer residual network (ResNet50) as base models for \texttt{HAM10000} and \texttt{Galaxy-Zoo} respectively. For \texttt{HateSpeech}, we use a $100$-dimensional \textit{fasttext} \citep{DBLP:journals/corr/JoulinGBDJM16} representation of each input and a ConvNet \citep{kim-2014-convolutional} as the base model.  We refer the reader to \citet{verma2022calibrated} for more details on the training and hyperparameter selection, as we follow their setup.  

\paragraph{Experimental Setup} We train the systems with an increasing number of experts, ranging from $2$ to $10$.  See Appendix \ref{sec:experimental_setup} for the details of how we simulate the experts. For each run, we enlarge the pool by adding increasingly accurate experts, and this process is repeated $3$ times with different random seeds.  We keep the base model fixed across these runs except for the additional output dimensions required by the expanded expert pool.  Ideally, the L2D systems should exhibit strictly increasing accuracy due to adding experts of increasing quality.  We compare our models against three baselines: one classifier, the best expert, and \citet{hemmer2022forming}'s MoE.

\paragraph{Results} The top row of Figure \ref{fig:MoE_comparison} reports the mean and standard error of the system accuracy as the number of experts increases.  While the OvA, softmax, and MoE models perform comparably on \texttt{HateSpeech} (left), OvA's performance (blue) is notably better on \texttt{HAM10000} (center) and \texttt{Galaxy-Zoo} (right) as its accuracy never falls below the one classifier baseline (red), while the others' accuracies do.

\subsection{Confidence Calibration}\label{sec: calibration}
In Section \ref{sec:confidence-calibration-theory-section}, we found that the two surrogates have very different estimators of $\mathbb{P}\left(\rsm_{j} = \ry_{i}\vert \rvx_{i}\right)$, the probability that the $j$th expert is correct.  We now test if these theoretical differences have consequences for practice.  To ensure ECE is well-defined for the softmax loss, we cap any confidences greater than $1$ at $1$.  In addition to reporting calibration for the preceding experiment (system accuracy), we also perform simulations using the standard splits of \texttt{CIFAR-10} \citep{krizhevsky2009learning}.  We use a $28$-layer wide residual network \citep{zagoruyko2016wide}, following \citet{verma2022calibrated}.

\paragraph{Simulation \#1: Increasing Experts} We perform a simulation to see how the methods perform under an increasing number of experts.  We generate a synthetic expert with a correctness probability of $70\%$ over the first five classes and random across all other classes.  We then replicate that expert and add it to the expert pool, ranging from $4$ to $20$ total experts.  Figure \ref{fig:avg-ECE-increase-experts-CIFAR-10} reports the average ECE across experts as the pool increases.  The OvA method (blue) is roughly stable at about $2\%$ ECE as experts are added.  The softmax method (orange) has roughly double the ECE ($\sim 4.5\%$). In Figure \ref{fig:acc-increase-experts-CIFAR-10}, we report the overall system accuracy to see if these calibration differences have an effect.  We see some positive effects, with OvA (blue) having a better accuracy for $12$ and fewer experts.  However, the softmax (orange) has the best accuracy at $20$ experts, despite its calibration still being worse.


\paragraph{Simulation \#2: Expert Dependence} We next perform a simulation to see how calibration error can propagate across the estimators.  We simulate four experts with one always being random and the other three having a probability of correctness that increases from $20\%$ to $95\%$ on the first five classes (random for others). We hypothesize that the softmax's ECE for the random expert will \emph{increase} when the probability of correctness for the other three experts increases due to the tied parameterization.  Figures \ref{fig:avg-ECE-increase-confidence-CIFAR-10} and \ref{fig:randomexpert-ECE-increase-confidence-CIFAR-10} report the results, with the former reporting average ECE and the latter the ECE of just the random expert.  Firstly, from Figure \ref{fig:avg-ECE-increase-confidence-CIFAR-10}, we see that again the OvA method is better calibrated across all experimental settings.  Then from Figure \ref{fig:randomexpert-ECE-increase-confidence-CIFAR-10}, we see that our hypothesis is confirmed: OvA (blue) is able to model the random expert well no matter the other experts' abilities, but the softmax (orange) is not.  The softmax's ECE increases almost in-step with the expert correctness, except for some cancellation effect happening at $80\%$.  This is clearly an undesirable behavior from the standpoint of safety since any ECE above zero means that the system is reporting that the expert is better than random and thus misleading the user.

\paragraph{Hate Speech, HAM10000, and Galaxy-Zoo}  Lastly we report the calibration error of the models trained for the experiments reported in Section \ref{sec:system-accuracy}.  The results are in the bottom row of Figure \ref{fig:MoE_comparison}.  The trend we observe in the \texttt{CIFAR-10} simulations is also observed here, with OvA (blue) having the best calibration.  This may explain why OvA has the best system accuracy.  Unsurprisingly, the MoE has extremely poor calibration, which is likely due to its inconsistent optimization objective which allows for sub-optimal models (as we prove in Proposition \ref{prop:moe_inconsis}).



\subsection{Conformal Ensembles}\label{sec:ensemble_results}
Lastly, we study our proposal of using CI to ensemble multiple experts.  We first analyze the two proposed statistics, demonstrating the regularized version's superior ability to recover the experts who are oracles.  We then report the downstream effect on the overall system accuracy, comparing performance to that of a fixed-size ensemble of experts.  


\paragraph{Experts and Setup} We experiment with two settings on \texttt{CIFAR-10}, each with $10$ total experts.  In the first (\textit{no noise}), we synthesize experts such that they are an oracle on an increasing subset of the classes and guaranteed to be wrong on the classes not in that set.  In the second (\textit{with noise}), the experts are oracles in the same way but now have a (uniformly) random chance of being correct for the non-oracle classes.  The theory of CI guarantees that the sets \textit{marginally} cover all oracle experts.  Yet, ideally, we wish the sets to contain \emph{only} the experts who are oracles.  We use $\alpha=0.1$ in all experiments.


\begin{figure*}
     \centering


     \begin{subfigure}[b]{0.325\textwidth}
         \centering
         \includegraphics[width=\linewidth]{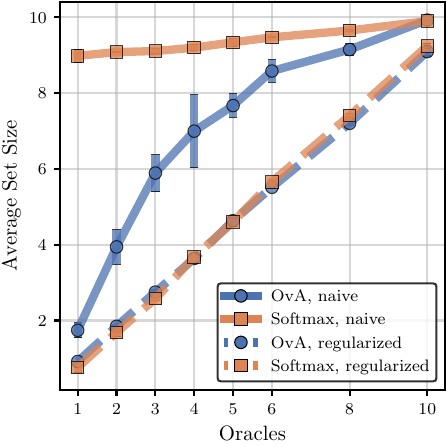}
         \caption{Set Size, No Noise}
         \label{fig:non-randomized-set-size-CIFAR-10}
     \end{subfigure}
     \hfill
    \begin{subfigure}[b]{0.325\textwidth}
         \centering
         \includegraphics[width=\linewidth]{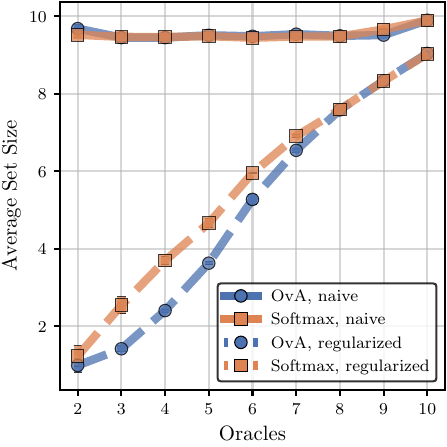}
         \caption{Set Size, with Noise}
         \label{fig:randomized-set-size-CIFAR-10}
     \end{subfigure}
     \hfill
     \begin{subfigure}[b]{0.325\textwidth}
         \centering
         \includegraphics[width=\linewidth]{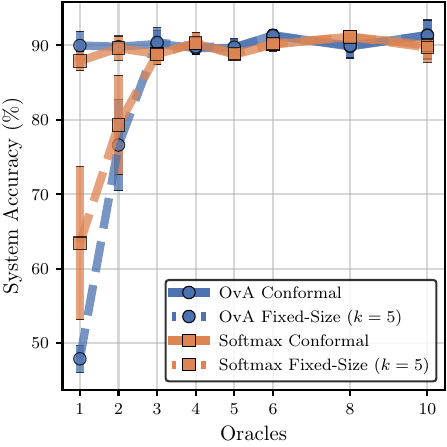}
         \caption{CIFAR-10: Sys.~Accuracy}
         \label{fig:system-accuracy-ensemble-CIFAR-10}
     \end{subfigure}
     \label{}
     \caption{\textit{Conformal Sets of Experts.} The figures above report our analysis of the two statistics proposed in Section 5.  Subfigures (a) and (b) show the ability of the two statistics to select the correct number of experts as the number of oracle experts increases---so optimal performance is the $y=x$ line.  Subfigure (a) reports the no-noise setting (so experts are either perfectly incorrect or correct), and we see that the naive statistic (solid lines) overestimates the set size.  The problem is even worse in Subfigure (b).  However, the regularized statistic (dashed lines) is able to do well in both cases.  Subfigure (c) shows how ensembling the set affects system accuracy.  The conformal approach is able to out-perform a fixed size of $5$ experts for a small number of oracles and is equivalent at higher numbers.}
\end{figure*}

\paragraph{Expert Identification}  The CI results are reported in Figures \ref{fig:non-randomized-set-size-CIFAR-10} and \ref{fig:randomized-set-size-CIFAR-10}.  The number of oracles is on the y-axis, and the average set size is on the x-axis.  Optimal performance would be the $y=x$ line.  The results for the no-noise setting are reported in Figure \ref{fig:non-randomized-set-size-CIFAR-10}.  The naive statistic (solid lines) considerably inflates the set size for both softmax and OvA.  Yet OvA is much closer to $y=x$, which suggests superior calibration leads to better CI.  The performance of the regularized statistic is shown by the dashed lines.  Both softmax and OvA perform nearly perfectly.  Figure \ref{fig:randomized-set-size-CIFAR-10} reports the with-noise setting, and we find that the naive statistic performs terribly for both losses.  The regularized statistic, on the other hand, performs well for both softmax and OvA.  Softmax demonstrates slight superiority for $2-5$ oracles.


\paragraph{Overall System Accuracy}  We next investigate using the conformal set as an ensembling strategy.  Upon deferral, we use majority voting across the set to generate the final prediction.  We compare this with the baseline of using a fixed ensemble size---specifically, the top five ranked experts.  We show the results for \texttt{CIFAR-10} in Figure \ref{fig:system-accuracy-ensemble-CIFAR-10}.  The crucial settings are for one and two oracles since using a fixed ensemble size is guaranteed to fail---as is confirmed by the plot ($<80\%$ accuracy).  We see that the conformal ensembles are clearly superior here, achieving around $90\%$ accuracy.  For three or more oracles, both methods have equal performance.  This is expected since only three oracles are needed to form a correct majority.  
We emphasize that CI's \textit{adaptivity} is highly desirable so that the best experts can be identified with transparency and queried efficiently.




\section{CONCLUSIONS}
We have extended the L2D framework to support multiple experts.  We proposed two optimization objectives and proved that they are both consistent.  Our proposed optimization objectives are simple to use in practice and could be embedded into any empirical risk minimization framework. Additionally, we also studied their potential to be confidence calibrated, showing that the softmax-based objective can result in mis-calibrated models in practice.  Lastly, we considered a principled procedure for selecting \emph{minimal} sets of experts to ensemble.  For future work, we aim to improve the data efficiency of our method by extending the active learning results of \citet{charusaie2022sample} to the multi-expert setting.  Additionally, recent work by \citet{narasimhan2022posthoc} has shown that L2D models can be prone to under-fitting when querying experts incurs additional cost and propose a post-hoc correction for surrogate losses.  Extending this methodology to the multi-expert setting may generally improve our results.

\subsubsection*{Acknowledgements}
This publication is part of the project \textit{Continual Learning under Human Guidance} (VI.Veni.212.203), which is financed by the Dutch Research Council (NWO). This work has also been partly supported by the Spanish government (AEI/MCI) under grant PID2021-123182OB-I00, by Comunidad de Madrid under grants IND2022/TIC-23550 and IntCARE-CM, and by the European Union (FEDER) and the European Research Council (ERC) through the European Union’s Horizon 2020 research and innovation program under grant 714161. The work by Daniel Barrejon has been additionally funded by the Spanish Ministerio de Educaci\`on, Cultura Deporte, grant FPU19/02681. This work was carried out on the Dutch national e-infrastructure with the support of the SURF Cooperative. 

\bibliography{references}
\bibliographystyle{plainnat}

\newpage
\appendix
\onecolumn

\section{Proofs and Derivations}
\looseness=-1In this section, we provide proofs for the main results in the paper. We derive the Bayes optimal rule for L2D to multiple experts, and show that the surrogate losses proposed in the paper are consistent. Next, we show that the mixture of experts formulation \citep{hemmer2022forming} is not consistent. We continue the notation from the main paper. For simplicity, we do not worry about measure-theoretic considerations and assume that appropriate conditions hold that allow us to interchange summations and integrals, for example. We begin by giving a formal definition of what it means for a surrogate loss to be consistent.

\begin{definition}
(Consistent Loss Function). A surrogate loss function $\psi: \mathcal{C} \times \mathcal{Y} \rightarrow \mathbb{R}_{+}$ operating in the surrogate space $\mathcal{C} \subseteq \mathbb{R}^{K}$ along with some suitable decoding function $g: \mathcal{C} \rightarrow \mathcal{Y}$ is said to be consistent if for all distributions $\mathcal{D}$, $\forall \epsilon > 0$, $\exists\;\delta >0$ such that if 
\begin{align}\label{eqn:delta-minimization}
 \vert\mathbb{E}_{(\rvx, \ry) \sim \mathcal{D}}\left[\psi\left(\ry, 
\mathbf{c}(\rvx)\right)\right] - \inf_{\bf{u} \in \mathcal{C}}\mathbb{E}_{(\rvx, \ry) \sim \mathcal{D}}\left[\psi\left(\ry, \mathbf{u}\right)\right]\vert < \delta,
\end{align}

holds for a prediction function $h: \mathcal{X} \rightarrow \mathcal{C}$, $h(\rvx):= \mathbf{c}(\rvx) \in \mathbb{R}^{K}$, then it must hold that 
\begin{align}\label{eqn:epsilon-conv}
\mathbb{P}_{(\rvx, \ry)\sim \mathcal{D}}\left[g \circ h (\rvx) \neq \ry\right]\leq \mathbb{P}_{(\rvx, \ry) \sim \mathcal{D}}\left[h^{*}(\rvx) \neq \ry\right] + \epsilon,
\end{align} 
where $h^{*}:\mathcal{X} \rightarrow \mathcal{Y}$ is the Bayes optimal predictor.
\end{definition}
\looseness=-1A common example of one decoding function in machine learning is the $\argmax$ function. Intuitively, consistency implies that the minimization of a surrogate loss $\psi(\cdot)$ results in a prediction function $h(\cdot)$ whose expected error converges to the Bayes risk. 

\subsection{Bayes Rule for Learning to Defer with Multiple Experts}\label{sec:Bayes-rule-multiL2D-derivation}

We have $J$ experts and a classifier, where the system either allows the classifier to make the final prediction or defers to one of the $J$ experts. When the classifier makes the prediction, the system incurs loss $\ell_{\text{clf}}(\hat{\ry}, \ry)$ where $\hat{\ry} = h(\vx)$. When the system defers to the $j^{\text{th}}$ expert, it incurs a loss $\ell_{\text{exp}}(\rsm_{j}, \ry)$. In what follows, we frame the learning to defer problem as a general classification problem, and aim to find a function $g: \mathcal{X} \rightarrow \hat{\mathcal{Y}} := \mathcal{Y} \cup \{\bot_{1}, \bot_{2}, \ldots, \bot_{J}\}$ and $\vert\hat{\mathcal{Y}}\vert = K+J$ with the minimum expected loss (also known as \textit{risk}). We consider $g$ as modeling a probabilistic decision rule $\mathbf{\delta}(\hat{y}\vert \vx) := [\delta(\hat{y}=1\vert \vx), \delta(\hat{y}=2\vert \vx), \ldots, \delta(\hat{y}=K+J\vert \vx)]$ where $\delta(\hat{y}=i\vert \vx)$ denotes the confidence in making the $i^{\text{th}}$ decision for $\vx \sim \rvx$. We write \textit{risk} as follows:

\begin{align}\label{eq:L2D-risk}\begin{split}
  \mathcal{R}_{\mathcal{D}}[\mathbf{\delta}\left(\hat{y}\vert \rvx\right)]  = \mathlarger{\mathlarger{\sum}}_{i=1}^{\text{K}}\mathlarger{\mathlarger{\sum}}_{j=1}^{\text{K}+\text{J}}\bigintsss_{\rvx}\delta(\hat{y}=j\vert\rvx)\ell\left(\hat{\ry}=j, \ry=i\right)\mathbb{P}\left(\ry=i\right)\mathbb{P}\left(\rvx \vert \ry=i\right)d\rvx,
\end{split}
\end{align}
where $\ell: (\hat{\ry}, \ry) \mapsto \mathbb{R}_{+}$ is a general loss function, $i$ runs over the input label space, and $j$ runs over the output prediction space (classifier and all the experts). We further expand the risk in Equation \ref{eq:L2D-risk} based on the definition of the loss function in learning to defer
\begin{align*}\begin{split}\label{eq:L2D-risk}
  \mathcal{R}_{\mathcal{D}}[\mathbf{\delta}\left(\hat{y}\vert \rvx\right)]  &= \mathlarger{\mathlarger{\sum}}_{i=1}^{\text{K}}\bigintss_{\rvx}\Biggr(\mathlarger{\mathlarger{\sum}}_{j=1}^{\text{K}}\delta(\hat{y}_{j}\vert\rvx)\ell_{\text{clf}}\left(j, i\right) \\ & +  \mathlarger{\mathlarger{\sum}}_{j=\text{K}+1}^{\text{K}+\text{J}}\left(\mathlarger{\mathlarger{\sum}}_{m=1}^{\text{K}}\delta(\hat{y}_{j}\vert\rvx)\ell_{exp}\left(m_{j}, \ry)\mathbb{P}(m_{j}\vert \rvx, \ry=i)\right) \right)\Biggr) \mathbb{P}\left(\ry=i\right)\mathbb{P}\left(\rvx \vert \ry=i\right)d\rvx,
\end{split}
\end{align*}
where we have used shorthand $\delta\left(\hat{y}_{j} \vert \rvx\right)$ to denote $\delta\left(\hat{y} = j \vert \rvx\right)$, and $\mathbb{P}\left(m_{j} \vert \rvx, \ry=i\right) = \mathbb{P}\left(\rsm_{j} = m \vert \rvx, \ry=i\right)$.
Next, we denote:
\begin{align*}
\begin{split}
w_{i,j} &= \ell_{\text{clf}}(j,i) \\
w_{i, \bot_{j}} &= \mathlarger{\mathlarger{\sum}}_{m=1}^{\text{K}}\delta(\hat{y}_{j}\vert\rvx)\ell_{exp}\left(m_{j}, \ry)\mathbb{P}(m_{j}\vert \rvx, \ry=i)\right)
\end{split}
\end{align*}
Thus, $\mathcal{R}_{\mathcal{D}}[\mathbf{\delta}\left(\hat{y}\vert \rvx\right)]$ can be written as:
\begin{align*}
    \mathcal{R}_{\mathcal{D}}[\mathbf{\delta}\left(\hat{y}\vert \rvx\right)] &= \mathlarger{\mathlarger{\sum}}_{i=1}^{\text{K}}\bigintsss_{\rvx}\left(\mathlarger{\mathlarger{\sum}}_{j=1}^{\text{K}}\delta(\hat{y}_{j}\vert\rvx)w_{i,j} + \mathlarger{\mathlarger{\sum}}_{j=\text{K}+1}^{\text{K}+\text{J}}\delta(\hat{y}_{j}\vert\rvx)w_{i, \bot_{j}}\right)\mathbb{P}\left(\ry=i\right)\mathbb{P}\left(\rvx \vert \ry=i\right)d\rvx.
\end{align*}
Denote $w^{*}_{i, \bot}:= \min_{j \in [J]}\{w_{i, \bot_{j}}\}$, we have
\begin{align*}
    \begin{split}
         \mathcal{R}_{\mathcal{D}}[\mathbf{\delta}\left(\hat{y}\vert \rvx\right)] \geq \mathlarger{\mathlarger{\sum}}_{i=1}^{\text{K}}\bigintsss_{\rvx}\left(\mathlarger{\mathlarger{\sum}}_{j=1}^{\text{K}}\delta(\hat{y}_{j}\vert\rvx)w_{i,j} + \mathlarger{\mathlarger{\sum}}_{j=\text{K}+1}^{\text{K}+\text{J}}\delta(\hat{y}_{j}\vert\rvx)w^{*}_{i, \bot}\right)\mathbb{P}\left(\ry=i\right)\mathbb{P}\left(\rvx \vert \ry=i\right)d\rvx.
    \end{split}
\end{align*}
We also denote $\sum_{j=\text{K}+1}^{\text{K} + \text{J}}\delta(\hat{y}_{j}\vert \rvx)$ as $\delta(\hat{y}_{\bot}\vert \rvx)$. Then the lower bound of $\mathcal{R}_{\mathcal{D}}[\mathbf{\delta}\left(\hat{y}\vert \rvx\right)]$, denoted as $\bar{\mathcal{R}}_{\mathcal{D}}[\mathbf{\delta}\left(\hat{y}\vert \rvx\right)]$, is
\begin{align*}
    \bar{\mathcal{R}}_{\mathcal{D}}[\mathbf{\delta}\left(\hat{y}\vert \rvx\right)] = \mathlarger{\mathlarger{\sum}}_{i=1}^{\text{K}}\bigintsss_{\rvx}\left(\mathlarger{\mathlarger{\sum}}_{j=1}^{\text{K}}\delta(\hat{y}_{j}\vert\rvx)w_{i,j} + \delta(\hat{y}_{\bot}\vert\rvx)w_{i, \bot}^{*}\right)\mathbb{P}\left(\ry=i\right)\mathbb{P}\left(\rvx \vert \ry=i\right)d\rvx.
\end{align*}

Since $\sum_{j=1}^{\text{K}}\delta(\hat{y}_{j}\vert \rvx) + \delta(\hat{y}_{\bot}\vert \rvx) = 1.0$ and $\int_{\rvx}\mathbb{P}(\rvx|\ry=i)d\rvx = 1.0$, we follow \citet{Chow1957AnOC} to decompose 
$\bar{\mathcal{R}}_{\mathcal{D}}[\mathbf{\delta}\left(\hat{y}\vert \rvx\right)]$ in two terms:
\[\bar{\mathcal{R}}_{\mathcal{D}} = \bar{\mathcal{R}}_{\mathcal{D}}^{\bot} + \bar{\mathcal{R}}_{\mathcal{D}}^{\mathbf{\delta}}, \]
where
\begin{align*}
    \begin{split}
        \bar{\mathcal{R}}_{\mathcal{D}}^{\bot} &= \sum_{i=1}^{\text{K}}\mathbb{P}(\ry=i)\cdot w^{*}_{i, \bot}, \\
        \bar{\mathcal{R}}_{\mathcal{D}}^{\mathbf{\delta}} &= \bigintss_{\rvx}\sum_{j=1}^{[\text{K}] \cup \{\bot\}}\delta(\hat{y}_{j}\vert \rvx) Z_{j}(\rvx) d\rvx, \text{and} \\
        Z_{j}(\rvx) &= \sum_{i=1}^{\text{K}}\left(w_{i,j} - w^{*}_{i,\bot}\right)\mathbb{P}(\rvx)\mathbb{P}(\ry=i\vert \rvx), \; j \in \{1,2, \ldots, \text{K}, \bot\}.
    \end{split}
\end{align*}
To elaborate, we simplify the problem from deferring to multiple experts to deferring to just the one expert with the minimum $w_{i, \bot_{j}}$ in obtaining the lower bound $\bar{\mathcal{R}}_{\mathcal{D}}^{\bot}$. We also observe that we have no control over $\bar{\mathcal{R}}_{\mathcal{D}}^{\bot}$. However, we can control $\bar{\mathcal{R}}_{\mathcal{D}}^{\mathbf{\delta}}$ by controlling the decision rule $\mathbf{\delta}$. We have $Z_{\bot}(\rvx) = 0$, and also it holds that
\begin{align*}
    \bar{\mathcal{R}}_{\mathcal{D}}^{\mathbf{\delta}} \geq \int_{\rvx}\min_{j} \left[Z_{j}\left(\rvx\right)\right]d\rvx,
\end{align*}
where the equality holds iff $\delta\left(\hat{y}_{k}\vert \rvx\right) = 1.0$ for $k = \argmin_{j}Z_{j}\left(\rvx\right)$. Thus, the optimal rule is to deterministically (i.e. with confidence 1.0) choose $k \in \{1,2, \ldots, \text{K}, \bot\}$ with the minimum $Z_{j}\left(\rvx\right)$.
This means that choosing $j$ for which the $Z_{j}(\rvx)$ is the smallest minimizes the risk. Given that $Z_{\bot} = 0$, this means that the classifier predicts when the minimum $Z_{j}\left(\rvx\right)$ is negative. Thus, deferral happens when $Z_{j}(\rvx)$ is positive for all $j$, i.e., the optimal rejection rule $r^{*}(\rvx)$ is: 
 \begin{align}\label{eq:optimal-rejection-rule-MultiL2D}
     r^{*}(\rvx) &= \mathbb{I}\left[Z_{j}(\rvx) \geq 0; \forall j \in \{1,\ldots, \text{K}\}\right].
 \end{align}
 This rejection rule is similar to the learning to defer to one expert. Given the definition of $Z_{j}(\rvx)$, the optimal behavior to choose which expert to defer to is the one with minimum $w_{i,\bot{j}}$. We further simplify the optimal rejection rule in the following proposition.
 
 \begin{proposition}
     The Bayes optimal rejection rule for L2D with multiple experts is given as:
     \begin{align}
         r^{*}(\rvx) = \begin{cases} 1 \;\; \text{if} \;\; \mathbb{E}_{\ry\vert\rvx}\left[\ell_{\text{clf}}(\hat{\ry}, \ry)\right] \geq \min_{j \in J}\mathbb{E}_{\ry\vert \rvx}\mathbb{E}_{\rsm\vert\rvx,\ry}\left[\ell_{\text{exp}}\left(\rsm_{j}, \ry\right)\right]\;\; \forall \hat{\ry} \in \mathcal{Y}\\ 
         0 \;\; \text{otherwise}.
         \end{cases}
     \end{align}
 \end{proposition}
 \textbf{Proof}:
The proof follows immediately from the definition of $r^{*}(\rvx)$ in Equation \ref{eq:optimal-rejection-rule-MultiL2D}.

In our work, we use the canonical $0$-$1$ loss for both $\ell_{\text{clf}}$ and $\ell_{\text{exp}}$. In this case, the rejection rule can trivially be written as in the following corollary.
\begin{corollary}
 For a misclassification $0$-$1$ loss, the optimal rejection rule is:
 \begin{align}
         r^{*}(\rvx) = \mathbb{I}\left[\max_{j \in J} \mathbb{P}(\rsm_{j} = \ry \vert \rvx = \vx) \geq \max_{y \in \mathcal{Y}}\mathbb{P}(\ry=y\vert \rvx=\vx)\right],\end{align} where
   $\mathbb{P}(\rsm_{j} = \ry \vert \rvx = \vx)$ is  the expert's correctness probability for the $j^{\text{th}}$ expert, and   $\mathbb{P}(\ry=y\vert \rvx=\vx)$ is the regular class probability. 
\end{corollary}

To sum it up, the Bayes optimal rule is to compare the confidences of the experts and the classifier, and follow whosoever has the highest confidence. The rule is analogous to the single expert setting proved in \citet{pmlr-v119-mozannar20b}.


\subsection{Proof of Theorem 3.1: Consistency of \texorpdfstring{$\rpsi_{\text{SM}}^{J}$}{Multi Expert Softmax Surrogate Loss}}
\label{sec:proof-theorem-3.1-main-text}
Convexity of $\rpsi_{\text{SM}}^{J}$ is immediately clear. We provide the proof for consistency below:

For simplicity, we denote 
\begin{equation}
    \begin{split}
       &  -\log \left( \frac{\exp\{g_{y}(\vx)\}}{\sum_{y' \in \mathcal{Y}^{\bot}} \exp\{g_{y'}(\vx)\} }\right) = \zeta_{y}(\vx), \\ &
        -\log \left( \frac{\exp\{g_{\bot,j}(\vx)\}}{\sum_{y' \in \mathcal{Y}^{\bot}} \exp\{g_{y'}(\vx)\} }\right) = \zeta_{\bot,j}(\vx).
    \end{split}
\end{equation}
Then, $\rpsi_{\text{SM}}^{J}$ can be written as:
\begin{equation}
    \rphi_{\text{SM}}^{J}\left(g_{1},\ldots, g_{K}, g_{\bot, 1}, \ldots, g_{\bot, J}; \vx, y, m_{1}, \ldots, m_{J}\right) = \zeta_{y}(\vx) + \sum_{j=1}^{J}\mathbb{I}[m_{j}=y]\cdot\zeta_{\bot,j}(\vx).
\end{equation}

We consider the pointwise risk $\mathcal{C}[\rpsi_{\text{SM}}^{J}]$ defined as:

\begin{equation}
    \mathcal{C}[\rpsi_{\text{SM}}^{J}] = \mathbb{E}_{\ry|\rvx = \vx}\mathbb{E}_{\rsm |\rvx=\vx, \ry=y}\left[\rpsi_{\text{SM}}^{J}(g_{1},\ldots, g_{K}, g_{\bot,1},\ldots, g_{\bot,J}; \vx, y, m_{1},\ldots, m_{J}) \right],
\end{equation}

where $\rsm \vert \rvx=\vx, \ry=y$ is a compact representation for each $\rsm_{j} \vert \rvx=\vx, \ry=y$. Our setup assumes that each $\rsm_{j}$ is independent. Denote $\eta_{y}(\vx) = \mathbb{P}(\ry=y\vert \rvx=\vx)$, we expand the expectations:
\begin{align*}
    \mathcal{C}[\rpsi_{\text{SM}}^{J}] & = \sum_{y \in \mathcal{Y}}\eta_{y}(\vx)\cdot\zeta_{y}(\vx) + \sum_{j=1}^{J}\left(\sum_{y \in \mathcal{Y}}\eta_{y}(\vx)\sum_{m_{j} \in \mathcal{M}}\mathbb{P}(\rsm_{j}=m_{j}\vert \rvx=\vx, \ry=y)\mathbb{I}[m_{j}=y]\cdot\zeta_{\bot,j}(\vx)\right).\\
    \mathcal{C}[\rpsi_{\text{SM}}^{J}] & = \sum_{y \in \mathcal{Y}}\eta_{y}(\vx)\cdot\zeta_{y}(\vx) + \sum_{j=1}^{J}\left(\sum_{y \in \mathcal{Y}}\eta_{y}(\vx)\sum_{m_{j} \in \mathcal{M}}\mathbb{P}(\rsm_{j}=y\vert \rvx=\vx, \ry=y)\cdot\zeta_{\bot,j}(\vx)\right).\\
    \mathcal{C}[\rpsi_{\text{SM}}^{J}] & = \sum_{y \in \mathcal{Y}}\eta_{y}(\vx)\cdot\zeta_{y}(\vx) + \sum_{j=1}^{J}\mathbb{P}(\rsm_{j}=\ry\vert \rvx=\vx)\cdot\zeta_{\bot,j}(\vx).
\end{align*}

Next, we consider the minimizer of $\mathcal{C}[\rpsi_{\text{SM}}^{J}]$. Since we have established convexity, we can analyze the minimizers of $\mathcal{C}[\rpsi_{\text{SM}}^{J}]$ by taking the partial derivatives w.r.t. $g_{y}\{\vx\}$ and $g_{\bot, j}\{\vx\}$ respectively and set them to $0$.

Thus, w.r.t. $g_{y}\{\vx\}$, we have
\begin{equation}\label{eqn:softmax-bayes-classifier-condition}
\begin{split}
    \frac{\partial \mathcal{C}[\rpsi_{\text{SM}}^{J}]}{\partial g_{y}\{\vx\}} = 0  \implies \frac{\exp\{g_{y}(\vx)\}}{\sum_{y' \in \mathcal{Y}^{\bot}} \exp\{g_{y'}(\vx)\}} = \frac{\mathbb{P}(\ry=y\vert\rvx=x)}{1 - \sum_{j=1}^{J}\mathbb{P}(\rsm_{j}=\ry\vert\rvx=\vx)}.
\end{split}
\end{equation}
Similarly, w.r.t. $g_{\bot, j}\{\vx\}$ we have 
\begin{align}\label{eqn:softmax-bayes-rejector-condition}
\frac{\partial \mathcal{C}[\rpsi_{\text{SM}}^{J}]}{\partial g_{\bot, j}\{\vx\}} = 0 \implies 
\frac{\exp\{g_{\bot, j}(\vx)\}}{\sum_{y' \in \mathcal{Y}^{\bot}} \exp\{g_{y'}(\vx)\}} = \frac{\mathbb{P}(\rsm_{j}=\ry \vert \rvx=\vx)}{1 -  \sum_{j=1}^{J}\mathbb{P}(\rsm_{j}=\ry\vert\rvx=\vx)}.
\end{align}

The above equations hold true for optimal classifier and the rejector. Thus, if we take the decision as in the main text, we are agreeing with the Bayes solution (considering that denominators are same in both the above conditions).

\subsection{Proof of Theorem 3.2: Consistency of \texorpdfstring{$\rpsi_{\text{OvA}}^{J}$}{OvA Surrogate Loss}}\label{sec:proof-theorem-3.2-main-text}
The proof follows directly from \citet{verma2022calibrated}. However, for completion we provide the full proof below.

For the surrogate prediction functions $g_{1},\ldots, g_{K}, g_{\bot,1}, \ldots, g_{\bot,J}$, and the binary classification surrogate loss $\rphi : \{-1, 1\} \times \mathbb{R} \rightarrow \mathbb{R}_{+}$, $\rpsi_{\text{OvA}}^{J}$ takes the following pointwise-form:
\begin{equation*} 
\begin{split}
      \mathcal{C}[\rpsi_{\text{OvA}}^{J}] & = \rpsi_{\text{OvA}}^{J}(g_{1},\ldots, g_{K}, g_{\bot,1},\ldots, g_{\bot,J}; \vx, y, m_{1},\ldots, m_{J}) \\
      & = \rphi[g_{y}(\vx)] + \sum_{y' \in \mathcal{Y}, y' \ne y} \rphi[-g_{y'}(\vx)] \ \ + \sum_{j=1}^{J} \rphi[-g_{\bot,j}(\vx)] + \sum_{j=1}^{J} \mathbb{I}[m_{j}=y]\left(\rphi[g_{\bot,j}(\vx)] - \rphi[-g_{\bot,j}(\vx)]\right).
\end{split}
\end{equation*}

We consider the pointwise \textit{inner} $\rpsi_{\text{OvA}}$\textit{risk} for some $\rvx = \vx$ written as follows:
\begin{align*}
    \mathcal{C}[\rpsi_{\text{OvA}}^{J}] = \mathbb{E}_{\ry|\rvx = \vx}\mathbb{E}_{\rsm |\rvx=\vx, \ry=y}\left[\rpsi_{\text{OvA}}^{J}(g_{1},\ldots, g_{K}, g_{\bot,1},\ldots, g_{\bot,J}; \vx, y, m_{1},\ldots, m_{J}) \right],
\end{align*}

We expand both the expectations one-by-one below:
\begin{equation*}
\begin{split}
    \mathcal{C}[\rpsi_{\text{OvA}}^{J}] &= \mathbb{E}_{\ry|\rvx = \vx}\Biggr[\rphi[g_{y}(\vx)] + \sum_{y' \in \mathcal{Y}, y' \ne y} \rphi[-g_{y'}(\vx)] \ \ + \sum_{j=1}^{J} \rphi[-g_{\bot,j}(\vx)] \\ &+ \sum_{j=1}^{J}\left(\sum_{m_{j} \in \mathcal{M}}\mathbb{P}(\rsm_{j}=m_{j} \vert \rvx=\vx, \ry=y)\mathbb{I}\left[m_{j} = y \right]\left(\rphi[g_{\bot,j}(\vx)] - \rphi[-g_{\bot,j}(\vx)]\right)\right)\Biggr].
\end{split}
\end{equation*}

Denote $\mathbb{P}(\ry=y\vert \rvx=\vx)$ as $\eta_{y}(\vx)$, then
\begin{equation*}
\begin{aligned}
    \mathcal{C}[\rpsi_{\text{OvA}}^{J}] & = \mathbb{E}_{\ry|\rvx = \vx}\sum_{y \in \mathcal{Y}}\left[\rphi[g_{y}(\vx)] + \sum_{y' \in \mathcal{Y}, y' \ne y} \rphi[-g_{y'}(\vx)]\right]  \\ & + \sum_{j=1}^{J}\Bigg( \rphi[-g_{\bot,j}(\vx)] \\
    & + \sum_{y \in \mathcal{Y}}\eta_{y}(\vx)\Big[\sum_{m_{j} \in \mathcal{M}}\mathbb{P}(\rsm_{j}=m_{j} \vert \rvx=\vx, \ry=y)\mathbb{I}\left[m_{j} = y \right]\left(\rphi[g_{\bot,j}(\vx)] - \rphi[-g_{\bot,j}(\vx)]\right)\Big]\Bigg)\\
    \mathcal{C}[\rpsi_{\text{OvA}}^{J}] &= \sum_{y \in \mathcal{Y}}\eta_{y}(\vx)\left[\rphi[g_{y}(\vx)] + \sum_{y' \in \mathcal{Y}, y' \ne y} \rphi[-g_{y'}(\vx)]\right]  \\ &+ \sum_{j=1}^{J}\left( \rphi[-g_{\bot,j}(\vx)] + \underbrace{\sum_{y \in \mathcal{Y}}\eta_{y}(\vx) \sum_{m_{j} \in \mathcal{M}}\mathbb{P}(\rsm_{j}=y \vert \rvx=\vx, \ry=y)}_{\text{$\mathbb{P}\left(\rsm_{j} = \ry \vert \rvx=\vx\right)$}}\left(\rphi[g_{\bot,j}(\vx)] - \rphi[-g_{\bot,j}(\vx)]\right)\right)\\
\end{aligned}
\end{equation*}

\begin{align*}
\begin{split}
    \mathcal{C}[\rpsi_{\text{OvA}}^{J}] & = \sum_{y \in \mathcal{Y}}\eta_{y}(\vx)\left[\rphi[g_{y}(\vx)] + \sum_{y' \in \mathcal{Y}, y' \ne y} \rphi[-g_{y'}(\vx)]\right] \\
    & + \sum_{j=1}^{J}\bigg[\rphi[-g_{\bot,j}(\vx)] + \mathbb{P}\left(\rsm_{j} = \ry \vert \rvx=\vx\right)\left(\rphi[g_{\bot,j}(\vx)] - \rphi[-g_{\bot,j}(\vx)]\right)\bigg].
\end{split}
\end{align*}
Denote $\mathbb{P}(\rsm_{j} = \ry\vert \rvx=\vx)$ as $p_{\rsm_{j}}$, then we have
\begin{equation*}
\begin{split}
    \mathcal{C}[\rpsi_{\text{OvA}}^{J}] &= \sum_{y \in \mathcal{Y}}\eta_{y}(\vx)\left[\rphi[g_{y}(\vx)] + \sum_{y' \in \mathcal{Y}, y' \ne y} \rphi[-g_{y'}(\vx)]\right]  \\ &+ \sum_{j=1}^{J}\left[\left(1 - p_{\rsm_{j}}\right)\rphi[-g_{\bot,j}(\vx)] + p_{\rsm_{j}}\rphi[g_{\bot,j}(\vx)] \right]
\end{split}
\end{equation*}
We further simplify the above equation as follows:
\begin{align*}
\begin{split}
    \mathcal{C}[\rpsi_{\text{OvA}}^{J}] &= \sum_{y \in \mathcal{Y}}\left[\eta_{y}(\vx)\cdot\rphi[g_{y}(\vx)] + \left(1 - \eta_{y}(\vx)\right)\cdot \rphi[-g_{y}(\vx)]\right]  +  \sum_{j=1}^{J}\left[\left(1 - p_{\rsm_{j}}\right)\rphi[-g_{\bot,j}(\vx)] + p_{\rsm_{j}}\rphi[g_{\bot,j}(\vx)] \right].
\end{split}
\end{align*}

Thus, we conclude from the above expression that we  $\text{K}+\text{J}$ binary classification problems where the pointwise risk (or inner risk) for the $i^{th}$ binary classification problem is given as $\eta_{y}\left(\vx\right) \rphi\left(g_{y}\left(\vx\right)\right) + \left(1 - \eta_{y}\left(\vx\right)\right)\rphi\left(-g_{y}\left(\vx\right)\right)$ for $i \in [K]$ and $p_{\rsm_{j}}\left(\vx\right)\rphi\left(g_{\bot,j}\left(\vx\right)\right) + \left(1 - p_{\rsm_{j}}\left(\vx\right)\right) \rphi\left(-g_{\bot,j}\left(x\right)\right)$ when $i \in [J]\}$. Thus,  minimizer of the inner $\rpsi_{\text{OvA}}$ \textit{-risk} can be analyzed in terms of the pointwise minimizer of the \textit{inner} $\rphi$\textit{-risk} for each of the $\text{K}+\text{J}$ sub binary classification problems. Denote the minimizer of pointwise \textit{inner} $\rpsi_{\text{OvA}}$\textit{-risk} as $\boldsymbol{g}^{*}$, then the above decomposition means $g^{*}_{i}$ corresponds to the minimizer of the \textit{inner} $\rphi$\textit{-risk} for the $i$th binary classification problem.

We know that the Bayes solution for the binary classification problem is $\sign\left(\eta(\vx) - \frac{1}{2}\right)$ where $\eta(\vx)$ denotes $p(\ry=1|\rvx=\vx)$. Now when the binary surrogate loss $\rphi$ is a strictly proper composite loss for binary classification, by the property of strictly proper composite losses, we have $\sign(g^{*}_{y}(\vx))$ would agree with the Bayes solution of the Binary classification, i.e. $g^{*}_{y}(\vx) > 0$ if $\eta_{y}\left(\vx\right) > \frac{1}{2}$. And similarly $g^{*}_{\bot}\left(\vx\right) > 0$ if $p_{\rsm_{j}}\left(\vx\right) > \frac{1}{2}$. Furthermore, we have the existence of a continuous and increasing inverse link function $\rgamma^{-1}$ for the binary surrogate $\rphi$ with the property that $\rgamma^{-1}\left(g^{*}_{y}\left(\vx\right)\right)$ would converge to $\eta_{y}\left(\vx\right)$. Similarly, $\rgamma^{-1}\left(g^{*}_{\bot,j}\left(\vx\right)\right)$ would converge to $p_{\rsm_{j}}\left(\vx\right)$.

Thus, when the binary surrogate loss $\rphi$ is a strictly proper composite loss, and the classifier and the rejector are defined as in the main text, the minimizer of the pointwise risk $\mathcal{C}\left[\rpsi_{\text{OvA}}^{J}\right]$ agree with the Bayes optimal solution. Thus, $\rpsi_{\text{OvA}}^{J}$ is a calibrated loss function for L2D w.r.t. $0$-$1$ misclassification loss.

\subsection{Inconsistency of the  Mixture of Experts Formulation \texorpdfstring{\citep{hemmer2022forming}}{Hemmer et al. 2022}}

\begin{proposition}
\label{prop:moe_inconsis}
    $L_{\text{MoE}}$ is inconsistent for learning to defer.
\end{proposition}
The proof works by construction. Specifically, we construct a distribution over $\mathcal{X} \times \mathcal{Y}$ for which the necessary condition for consistency does not hold true.

Consider we have $\mathcal{X} = \{\vx\}$ and $\mathcal{Y} = \{0,1\}$, i.e. the input space contains the singleton element $\vx$ with $2$ output labels. We define the following distribution $\mathcal{D}$ such that $\mathbb{P}(\vx, 0) = \alpha_0$, $\mathbb{P}(\vx, 1) = \alpha_1$. For completion, $\alpha_0 + \alpha_1 = 1$. For simplicity, we consider one expert who predicts correctly with perfect confidence, i.e $\rsm = \ry$ $\forall y$.  The mixture of experts method works by estimating the allocator scores $w_e(\vx)$ (for expert) and $w_c(\vx)$ (for classifier) such that $w_e(\vx) + w_c(\vx) = 1$, and the classifier scores $c_0(\vx), c_1(\vx)$ with $\sum_{i=0}^{1}c_i(\vx) = 1$. In such a setting, we have 
\begin{align*}
    \begin{split}
        \mathbb{E}_{(\rvx, \ry) \sim \mathcal{D}}\left[L_{\text{MoE}}\left(F, A, \rvx, \ry, \rsm\right)\right] = -\alpha_{0}\left[\log\left(w_e + w_c \cdot c_0\right)\right] - \alpha_{1}\left[\log\left(w_e + w_c \cdot c_1\right)\right]. 
    \end{split}
\end{align*}
It is an easy argument to see that the minimum value of the above expression is 0, i.e. $\inf_{A, F} \mathbb{E}_{(\rvx, \ry) \sim \mathcal{D}}\left[L_{\text{MoE}}\left(F, A, \rvx, \ry, \rsm\right)\right] = 0$. 

MoE system decides to defer to the expert if $w_e(\vx) > w_c(\vx)$. For $\delta > 0$, choose $w_e(\vx) = 0.5 - \delta$ and $w_c(\vx) = 0.5 + \delta$. Note that $\forall \; \delta > 0$, the system would always decide not to defer to the expert. Also choose $c_0(\vx) = 1, c_1(\vx) = 0$. 
For such an allocator $\bar{A}$ and the classifier $\bar{F}$, 
\begin{align*}
    \begin{split}
        \mathbb{E}_{(\rvx, \ry) \sim \mathcal{D}}\left[L_{\text{MoE}}\left(\bar{F}, \bar{A}, \rvx, \ry, \rsm\right)\right] &= -\alpha_1\cdot\log\left(0.5 - \delta\right) \\
        &\leq -\alpha_1\cdot\left(0.5 - \delta - 1\right) = \alpha_1 \cdot\left(0.5 + \delta\right),
    \end{split}
\end{align*}
where the inequality comes from using $\log\left(x\right) \leq x - 1$. Next, choose $\alpha_1$ such that $\alpha_1 = \frac{\delta}{0.5 + \delta}$ (why this is true is left as an exercise to the reader). Combining everything, we have shown that
\[\vert \mathbb{E}_{(\rvx, \ry) \sim \mathcal{D}}\left[L_{\text{MoE}}\left(\bar{F}, \bar{A}, \rvx, \ry, \rsm\right)\right] - \inf_{A, F} \mathbb{E}_{(\rvx, \ry) \sim \mathcal{D}}\left[L_{\text{MoE}}\left(F, A, \rvx, \ry, \rsm\right)\right] \vert \leq \delta.\] 
Thus, our choice of $\bar{A}$ and $\bar{F}$ satisfy Equation \ref{eqn:delta-minimization} for all $\delta > 0$. Since in our construction, we always allow the decision to be made by the classifier which can only predict class label $h(\vx)$ $0$, we have $\mathbb{P}_{(\rvx, \ry) \sim 
\mathcal{D}}\left[h(\vx) \neq \ry\right] = \alpha_1$. And the Bayes optimal rule $h^{*}(\vx)$ is to always let the expert make the prediction, thus, $\mathbb{P}_{(\rvx, \ry) \sim \mathcal{D}}\left[h^{*}(\vx) \neq \ry\right] = 0$. Hence, we have 
\[
\mathbb{P}_{(\rvx, \ry) \sim \mathcal{D}}\left[h(\vx) \neq \ry\right] = \mathbb{P}_{(\rvx, \ry) \sim \mathcal{D}}\left[h^{*}(\vx) \neq \ry\right] +  \eta,
\]
where $\eta = \alpha_1$. Thus, $\forall \epsilon < \kappa, \epsilon > 0$, Equation \ref{eqn:epsilon-conv}
fails to hold true. Hence, we have shown that the optimization of $L_{\text{MoE}}$ allows faulty solutions that may not reach the Bayes optimal predictor.




\section{Choice of Hyperparameters for Regularized Conformal Ensembles}
\label{sec:ci_hypers}
We begin by giving a brief introduction to the procedure of conformal risk control. For detailed exposition, we refer the reader to the original paper \citep{angelopoulos2022conformal}. 

Conformal risk control \citep{angelopoulos2022conformal} is a generalized form of conformal prediction which aims to control any bounded monotone loss function $\ell(\cdot)$ in expectation. In our work, we are interested in False Negative Rate (FNR) as a specific loss function which satisfies the monotonicity property as a function of $\lambda$ (Equation \ref{eqn:conformal-risk-control}). Given access to the calibration data $\{\vx_n, y_n\}_{n=1}^{\text{N}}$, the goal in conformal risk control is to find $\hat{\lambda}$ so that the following coverage guarantee holds:
\[\mathbb{E}\left[\ell\left(C_{\hat{\lambda}}(\vx_{n+1}\right)\right] \leq \alpha.\]
Without loss of generality, we consider $\ell$ to be a non-increasing function of $\lambda$ and bounded by a constant $B$. Procedurally, it works by defining $S\left(\vx_{1:\text{N}}; \lambda \right) = \frac{1}{\text{N}}\sum_{i=1}^{\text{N}}\ell\left(C_{\lambda}\left(\vx_{i}\right)\right)$. For $\alpha \in (-\infty, B]$, $\hat{\lambda}$ is then defined as \[
\hat{\lambda} = \inf\left\{\lambda: \frac{n}{n+1}S\left(\vx_{1:\text{N}};\lambda\right) + \frac{B}{n+1} \leq \alpha\right\}.
\]
Assuming exchangeability on $\ell(C_{\lambda}(\vx_{i}))$, this results in the desired coverage guarantees for $C_{\hat{\lambda}}(\vx_{n+1})$. 

In our work, we use a grid of equally spaced $1500$ values in $[0,1]$ to pick $\hat{\lambda}$. We have two hyperparameters $\kappa$ and $\beta$ in the regularized conformal ensemble procedure discussed in Section 5.  Algorithm \ref{algo:choosing-beta} below details the procedure to choose $\kappa$. 

\SetAlgoNlRelativeSize{0}  
\begin{algorithm}
\caption{Choice of $\kappa$}
\label{algo:choosing-beta}
\KwData{Error rate: $\alpha$, $\#$ Experts: $\text{J}$, Data: $\{\mathbf{s}^{i}(\vx), \mathbf{e}^{i}(\vx)\}_{i=1}^{\text{N}}$ where $\mathbf{s}^{i}(\vx) \in [0,1]^{\text{J}}$ and $\mathbf{e}^{i}(\vx) \in \{0,1\}^{\text{J}}$}

$B \gets \Lbag \cdot \Rbag$\;

\For{i $\in$ $[1,N]$}
{
\For{j $\in$ $[1,J]$}{
\If{$\mathbb{I}\left\{e_{j}^{i}(\vx) == 1\right\}$}
{$B \gets B \cup \{s_{j}^{i}(\vx)\}$}
}
$S \gets \text{sort}\left(B\right)$ s.t. $u_{i}, u_{j} \in S, i \leq j$, then $u_{i} \geq u_{j}$\; 

$\kappa^{*} \gets 1-\alpha$ $\text{quantile}$ of $S$\;
}
\end{algorithm}
We can employ corrections to account for finite sample size $\text{N}$ on line $9$. Given this procedure to choose $\kappa^{*}$, one may argue that our choice of $\kappa^{*}$ can give us meaningful prediction sets by designing a prediction set as:
\[C_{2}(\vx) = \{j : s_{j}(\vx) \geq \kappa^{*}\}.\]
However, our next proposition establishes that $C_{\lambda}(\vx)$ results in prediction sets at most as large as $C_{2}(\vx)$.

\begin{proposition}
    Define the prediction sets $C_{\lambda}(\vx) = \{j : s_{j}(\vx) + \beta\left(s_{j}(\vx) - \kappa^{*}\right) > 1 - \lambda\}$ and $C_{2}(\vx) = \{j : s_{j}(\vx) \geq \kappa^{*}\}$, where $\kappa^{*}$ is defined as above, $\beta \geq 0$, $0 \leq \lambda \leq 1$, 
    then it trivially holds that
    \[C_{\lambda}(\vx) \subseteq C_{2}(\vx).\]
\end{proposition}
We tune $\beta$ in a grid-search manner. The grid size for $\beta$ is $[3.5, 1e^{-3}]$ with steps of $50$ samples.  We split the total number of deferred samples into two portions: one for tuning hyperparameters $\beta$ and $\kappa$, another for the regular conformal procedure. $30\%$ of the deferred samples are used to tune the ensembling hyperparameters.

\section{Experimental Setup for Simulated Experts}\label{sec:experimental_setup}
For the experiment regarding the overall system accuracy described in Section \ref{sec:system-accuracy} we simulate $10$ unique experts of increasing ability. Below you can find the description of the experts' configurations for the studied datasets. 

\subsection{Hate Speech and Galaxy-Zoo}
For \texttt{Galaxy-Zoo} and \texttt{HateSpeech}, we define the following experts using the human annotations available in the datasets and using various perturbations of these predictions:
\begin{enumerate}
    \item \textcolor{solin-colour2}{Human expert}: we sample predictions from the provided human annotations.
    \item \textcolor{solin-colour5}{Flipping human expert}: Expert who flips the given prediction with some probability \texttt{$\text{p}_\text{flip}$}.
    \item \textcolor{solin-colour4}{Probabilistic expert}: Expert who makes use of the annotations with some probability \texttt{$\text{p}_\text{annotator}$}, or predicts randomly otherwise.
\end{enumerate}
The whole expert configuration is described in Table \ref{tab:tab:hatespeech_expert_config}.

\begin{table}[H]
\caption{Hate Speech and Galaxy-Zoo experts configuration.}
\centering
\begin{tabular}{rlcc}
  \hline
  & Expert configuration & \texttt{$\text{p}_\text{flip}$}$[\%]$  & \texttt{$\text{p}_\text{annotator}$}$[\%]$ \\ 
  \hline
  1 & \textcolor{solin-colour7}{Random Expert} & - & - \\
  2 & \textcolor{solin-colour4}{Probabilistic Expert} & - & 10 \\
  3 & \textcolor{solin-colour5}{Flipping Human Expert} & 50 & - \\
  4 & \textcolor{solin-colour4}{Probabilistic Expert} & - & 75 \\
  5 & \textcolor{solin-colour5}{Flipping Human Expert} & 30 & - \\
  6 & \textcolor{solin-colour5}{Flipping Human Expert} & 20 & - \\
  7 & \textcolor{solin-colour4}{Probabilistic Expert} & - & 85 \\
  8 & \textcolor{solin-colour2}{Human Expert} & - & - \\
  9 & \textcolor{solin-colour4}{Probabilistic Expert} & - & 50 \\
  10 & \textcolor{solin-colour2}{Human Expert} & - & - \\
  \hline
\end{tabular}
\label{tab:tab:hatespeech_expert_config}
\end{table}

\subsection{HAM10000}

The \texttt{HAM10000} dataset \citep{tschandl2018ham10000} is composed of dermatoscopic images corresponding to 
7 diagnostic categories in the realm of pigmented lesions. These 7 categories can be further decomposed into \textcolor{solin-colour2}{benign}: melanocytic nevi (\texttt{nv}), benign keratinocytic lesions (\texttt{bkl}), dermatofisbromas (\texttt{df}) and vascular lesions (\texttt{vasc}); and \textcolor{solin-colour3}{malign}: melanomas (\texttt{mel}), basal cell carcinomas (\texttt{bcc}) and actinic keratoses and intraepithelial carcinomas (\texttt{akiec}).

In contrast to the Galaxy-Zoo and Hatespeech dataset, for HAM10000 we do not have individual annotators predictions, but just the ground truth label. Further information can be found in the original dataset description \citep{tschandl2018ham10000}. In order to recreate a setup comparable to a real-world scenario, we create different experts configurations: 
\begin{enumerate}
    \item \textbf{Random expert}: This expert predicts randomly among all classes. 
    \item \textbf{Dermatologist expert}: These experts will be specialized in a set of categories, and will predict with probability \texttt{$\text{p}_\text{in}$}. Out of that set, they will predict with probability \texttt{$\text{p}_\text{out}$}.
    \item \textbf{MLPMixer}: We derive \texttt{HAM10000}'s expert predictions from the predictions of an $8$-layer MLPMixer \citep{Tolstikhin2021MLPMixerAA}, which has access to additional metadata such as age, gender, and diagnosis type.
\end{enumerate}

As it can be seen in Table \ref{tab:ham10000_expert_config}, we gradually add experts from a random expert to a final expert which simulates an experienced dermatologist. From \cite{kittler2002diagnostic} we know that clinical diagnosis of cutaneous melanoma with the unaided eye is only about $60\%$ accuracy, and that dermatologists equipped with dermatoscope can achieve accuracies of $75\%-84\%$. That is the reason why we chose for the simulated dermatologist experts to have those probabilities \texttt{$\text{p}_\text{in}$} and \texttt{$\text{p}_\text{out}$}.

\begin{table}[h]
\caption{HAM10000 experts configuration.}
\centering
\begin{tabular}{rlccc}
  \hline
  & Expert configuration & \texttt{$\text{p}_\text{in}$} $[\%]$ & \texttt{$\text{p}_\text{out}$} $[\%]$ & Diagnostic Category [\texttt{in}]\\ 
  \hline
  1 & \textcolor{solin-colour7}{Random Expert} & - & - & [\textcolor{solin-colour2}{\texttt{nv}}, \textcolor{solin-colour2}{\texttt{bkl}}, \textcolor{solin-colour2}{\texttt{df}}, \textcolor{solin-colour2}{\texttt{vasc}}, \textcolor{solin-colour3}{\texttt{mel}}, \textcolor{solin-colour3}{\texttt{bcc}}, \textcolor{solin-colour3}{\texttt{akiec}}]\\
  2 & \textcolor{solin-colour5}{Dermatologist for \textcolor{solin-colour3}{malign}} & 25 & 15 & [\textcolor{solin-colour3}{\texttt{mel}}, \textcolor{solin-colour3}{\texttt{bcc}}, \textcolor{solin-colour3}{\texttt{akiec}}]\\
  3 & \textcolor{solin-colour5}{Dermatologist for \textcolor{solin-colour2}{benign}} & 25 & 15 & [\textcolor{solin-colour2}{\texttt{nv}}, \textcolor{solin-colour2}{\texttt{bkl}}, \textcolor{solin-colour2}{\texttt{df}}, \textcolor{solin-colour2}{\texttt{vasc}}] \\
  4 & \textcolor{solin-colour6}{Specialized dermatologist in \texttt{nv}} & 50 & 15 & [\textcolor{solin-colour2}{\texttt{nv}}]\\
  5 & \textcolor{solin-colour6}{Specialized dermatologist in \texttt{vasc}} & 70 & 15 & [\textcolor{solin-colour2}{\texttt{vasc}}]\\
  6 & \textcolor{solin-colour6}{Specialized dermatologist in \texttt{mel}} & 75 & 15 & [\textcolor{solin-colour3}{\texttt{mel}}]\\
  7 & \textcolor{solin-colour5}{Dermatologist for \textcolor{solin-colour2}{benign}} & 75 & 25 & [\textcolor{solin-colour2}{\texttt{nv}}, \textcolor{solin-colour2}{\texttt{bkl}}, \textcolor{solin-colour2}{\texttt{df}}, \textcolor{solin-colour2}{\texttt{vasc}}] \\
  8 & \textcolor{solin-colour4}{MLP Mixer} & - & - & [\textcolor{solin-colour2}{\texttt{nv}}, \textcolor{solin-colour2}{\texttt{bkl}}, \textcolor{solin-colour2}{\texttt{df}}, \textcolor{solin-colour2}{\texttt{vasc}}, \textcolor{solin-colour3}{\texttt{mel}}, \textcolor{solin-colour3}{\texttt{bcc}}, \textcolor{solin-colour3}{\texttt{akiec}}]\\
  9 & \textcolor{solin-colour1}{Experienced dermatologist} & 80 & 50 & [\textcolor{solin-colour2}{\texttt{nv}}, \textcolor{solin-colour2}{\texttt{bkl}}, \textcolor{solin-colour2}{\texttt{df}}, \textcolor{solin-colour2}{\texttt{vasc}}, \textcolor{solin-colour3}{\texttt{mel}}, \textcolor{solin-colour3}{\texttt{bcc}}, \textcolor{solin-colour3}{\texttt{akiec}}]\\
  10 & \textcolor{solin-colour1}{Experienced dermatologist} & 80 & 60 &  [\textcolor{solin-colour2}{\texttt{nv}}, \textcolor{solin-colour2}{\texttt{bkl}}, \textcolor{solin-colour2}{\texttt{df}}, \textcolor{solin-colour2}{\texttt{vasc}}, \textcolor{solin-colour3}{\texttt{mel}}, \textcolor{solin-colour3}{\texttt{bcc}}, \textcolor{solin-colour3}{\texttt{akiec}}]\\
  \hline
\end{tabular}
\label{tab:ham10000_expert_config}
\end{table}

\section{Additional Experiments and Results}

\subsection{Overlapping expertise among experts}
\begin{figure}
    \centering
    \begin{subfigure}[b]{0.3\textwidth}
         \centering
         \includegraphics[width=0.8\linewidth]{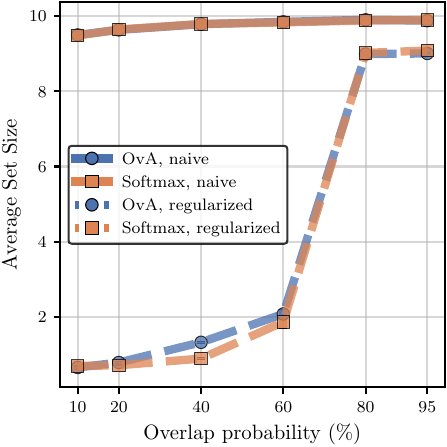}
         \caption{Set size.}
         \label{fig:set_size_overlap}
    \end{subfigure}
    \hfill
    \begin{subfigure}[b]{0.3\textwidth}
         \centering
         \includegraphics[width=0.8\linewidth]{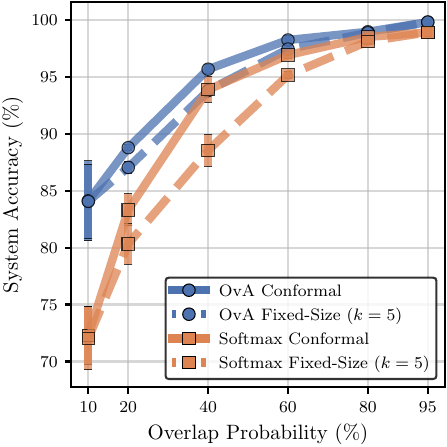}
         \caption{Sys.~Acc, naive conformal.}
         \label{fig:sys_acc_naive_overlap}
    \end{subfigure}
     \hfill
     \begin{subfigure}[b]{0.3\textwidth}
         \centering
         \includegraphics[width=0.8\linewidth]{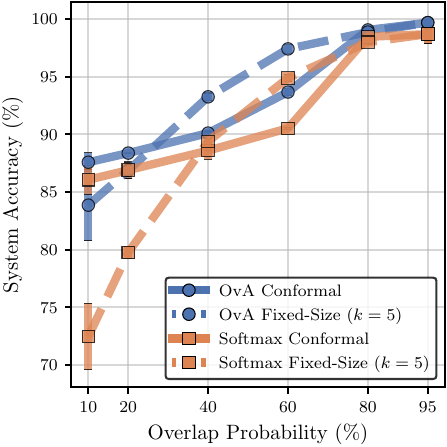}
         \caption{Sys.~Acc, regularized conformal.}
         \label{fig:sys_acc_reg_overlap}
     \end{subfigure}
     \caption{\textit{Gradual overlapping expertise results.} The figures above report the average set size (a) and system accuracies under the naive conformal method (b) and regularized conformal method (c) for increasing expertise overlap for \texttt{CIFAR-10}. From Figure (a) we see that the regularized conformal method is more dynamic than the naive method for both OvA and softmax. System accuracies for the naive conformal method (b) are slightly better than fixed-size ensemble because we ensemble all experts, whereas for the regularized conformal method (c) we see a slightly drop due to the adaptivity of the conformal sets}
     \label{fig:overlap_experiment}
\end{figure}

Similar to the conducted experiments in Section 7.3 in the main manuscript, we want to study ensembling multiple experts, this time under a different experiment setup. We will have 10 experts for the \texttt{CIFAR-10} dataset, each of them being an oracle on a specific class out of the $10$ classes from the dataset, and we will increase the overlapping probability of these experts being correct on the other classes where the experts are not oracle from $10\%$ to $95\%$ overlap. That is, we will vary from specialized experts to fully overlapped experts. We hope to see that the average set size for specialized experts is close to 1 and for fully overlapped experts close to the total number of experts. We report the results in  Figure \ref{fig:overlap_experiment}. 

\paragraph{Average Set Size} First of all, it is worth noticing from Figure \ref{fig:set_size_overlap} that the average set size for the naive conformal method, both for softmax and OvA, is always close to the total number of experts, for specialized and overlapped experts. If we remember, the \textbf{naive test statistic} is calculated among \textit{all correct experts}. This is a very important point, because if an expert happens to be correct outside of their expertise domain, this results in a very big non-conformity score because of the low confidence of such expert. That is, imagine for certain sample $\vx$ and class $y=3$, for low overlapping probabilities, we might have $E=3$, where $e=1$ could be the oracle for class $y=3$ and $e=2, e=3$ two experts that were correct by chance. From Equation 14 in the manuscript, we can expect that, best-case scenario $s_{\pi_1}>s_{\pi_2}>s_{\pi_3}$, or even worse $s_{\pi_1}>\dots>s_{\pi_8}>s_{\pi_2}>s_{\pi_3}$, because other experts' confidences could be also greater than confidences from correct experts by chance. Therefore, we will obtain bigger test-statistics that result in very larger conformal sets. This problem has already been addressed in \cite{angelopoulos2022conformal}.
However, notice how the \textbf{regularized test statistic} is capable of producing smaller sets. The idea is that now we optimize additional parameters (described in Section 5 in the manuscript) to ensure that confidences lower than a certain threshold are filtered out for the calculation of the test statistic. 

\paragraph{System Accuracies} In Figure \ref{fig:sys_acc_naive_overlap} and \ref{fig:sys_acc_reg_overlap} we report the system accuracies for the naive conformal method and the regularized conformal method respectively. For the \textbf{naive conformal method} we obtain better results than using a fixed-size ensemble of experts of size 5. Because set sizes are almost always close to the total number of experts, and we do majority voting, then as long as the is a correct expert plus an expert correct by chance, we will predict correctly. However, for the \textbf{regularized conformal method} we notice a drop in the system accuracy for lower overlapping probabilities. Since for such cases now the set sizes are smaller, we have smaller number of experts in the set and therefore less chance of having correct experts by chance in the ensemble. Despite this drop in the accuracy, we clearly have a more dynamic and less conservative ensembling method.

\vfill

\end{document}

%% file: math_commands.tex
\usepackage{amsmath,amsfonts,bm}
\usepackage{upgreek}

\def\eqref#1{equation~\ref{#1}}

\def\1{\bm{1}}

\def\rgamma{{\upgamma}}
\def\rphi{{\upphi}}
\def\rpsi{{\uppsi}}

\def\rsm{{\textnormal{m}}}

\def\ry{{\textnormal{y}}}

\def\rvw{{\mathbf{w}}}
\def\rvx{{\mathbf{x}}}

\def\vtheta{{\bm{\theta}}}

\def\vx{{\bm{x}}}

\DeclareMathAlphabet{\mathsfit}{\encodingdefault}{\sfdefault}{m}{sl}
\SetMathAlphabet{\mathsfit}{bold}{\encodingdefault}{\sfdefault}{bx}{n}

\DeclareMathOperator*{\argmax}{arg\,max}
\DeclareMathOperator*{\argmin}{arg\,min}

\DeclareMathOperator{\sign}{sign}